\documentclass[iicol,sn-mathphys,smallcondensed]{sn-jnl}

\usepackage{multirow}
\usepackage{array}
\usepackage{afterpage}
\usepackage{siunitx}
\usepackage{booktabs}
\usepackage{comment}

\title[Recent Techniques for Person Re-Id]{A Review of Recent Techniques for Person Re-Identification}

\author*[1]{\fnm{Andrea} \sur{Asperti}}\email{andrea.asperti@unibo.it}

\author[1]{\fnm{Salvatore} \sur{Fiorilla}}\email{salvatore.fiorilla@unibo.it}

\author[2]{\fnm{Simone} \sur{Nardi}}\email{simone.nardi@mermec-engineering.com}

\author[1]{\fnm{Lorenzo} \sur{Orsini}}\email{lorenzo.orsini@studio.unibo.it}

\affil*[1]{\orgdiv{Department of Informatics - Science and Engineering (DISI)}, \orgname{University of Bologna}, \orgaddress{\street{Mura Anteo Zamboni 7}, \city{Bologna}, \postcode{40126},
\country{Italy}
}}

\affil*[2]{\orgdiv{Department of Computational Science}, \orgname{Mermec Engineering srl}, \orgaddress{\street{Via Livornese 1019}, \city{San Piero a Grado (PI)}, \postcode{56122},
\country{Italy}
}}

\begin{document}

\abstract{
Person re-identification (ReId), a crucial task in surveillance, involves matching individuals across different camera views.
The advent of Deep Learning, especially supervised techniques like Convolutional Neural Networks and Attention Mechanisms, has significantly enhanced person Re-ID. However, the success of supervised approaches hinges on vast amounts of annotated data, posing scalability challenges in data labeling and computational costs. To address these limitations, recent research has shifted towards unsupervised person re-identification. Leveraging abundant unlabeled data, unsupervised methods aim to overcome the need for pairwise labelled data. Although traditionally trailing behind supervised approaches, unsupervised techniques have shown promising developments in recent years, signalling a narrowing performance gap.

Motivated by this evolving landscape, our survey pursues two primary objectives. First, we review and categorize significant publications in supervised person re-identification, providing an in-depth overview of the current state-of-the-art and emphasizing little room for further improvement in this domain. Second, we explore the latest advancements in unsupervised person re-identification over the past three years, offering insights into emerging trends and shedding light on the potential convergence of performance between supervised and unsupervised paradigms. This dual-focus survey aims to contribute to the evolving narrative of person re-identification, capturing both the mature landscape of supervised techniques and the promising outcomes in the realm of unsupervised learning.
}


\maketitle
\let\thefootnote\relax\footnotetext{\hspace*{-5mm}This footnote will be used only by the Editor and Associate Editors.~The edition in this area is not permitted to the authors. This footnote must not be removed while editing the manuscript.}




\section{INTRODUCTION}
In recent times, surveillance has expanded significantly with the proliferation of cameras in airports, stadiums, commercial centers, corporate buildings, and residential areas. These networks generate vast amounts of video data, traditionally monitored by humans, which is time-consuming and labor-intensive. As a result, the need arises to leverage computer vision and intelligent surveillance systems for efficient video analysis \cite{endtoend_reid, DLsurvey, nardi22}.

Within these intelligent, real-time video surveillance frameworks, Person Re-identification (Re-ID) has garnered considerable attention in recent years, due to its pivotal role and broad applications.  The problem consists in recognizing comparable subjects across a network of non-overlapping cameras, constituting what is commonly referred to as multi-camera surveillance systems \cite{LIP,tahir2021transformers,nardi23}. 

In its typical formulation, the Person Re-ID task involves identifying a specific individual from a large collection of person images, known as the gallery, using a query image. This task falls under the category of image retrieval problems, but the goal here is to determine the identity of the person in the query image, represented by a label called ID. The core task of person Re-ID consists in learning distinctive features that differentiate between images of the same person and those of different individuals. The challenge is further complicated by variations in viewpoint, pose, lighting, and image quality across different cameras in real-world scenarios, where individuals may appear in multiple cameras across various locations, making feature extraction more difficult~\cite{ming2022deep}.



Early research efforts in the realm of person re-identification included the utilization of approaches focused on hand-crafted feature extraction, by using methods such as colour histograms, texture features, scale-invariant feature transform (SIFT), local binary pattern (LBP), and other 
techniques~\cite{hand_craft}. Metric learning algorithms have also been employed, relying on distance metrics, support vector machines (SVMs), neural networks (NN), cross-view quadratic discriminant analysis (XQDA), and nearest neighbours (KNN). However, this approach necessitates expert proficiency in feature design and exhibits limitations in capturing complex relationships within images~\cite{hand_metric}. 

Recent advancements in Deep Learning have eliminated the need for manual feature extraction, as Deep Neural Networks can learn robust features directly from raw images. These methods now outperform traditional approaches and dominate research in this field \cite{eng_DL}, \cite{over_reid}. Their success stems from two factors: the complexity of network architectures, which effectively handle variations in posture, lighting, and background; and the ability to transform input images into highly semantic feature representations, enhancing model performance \cite{DL-strength}.


Thanks to the advancement of deep learning models, researchers have achieved increasingly promising outcomes, such as \cite{DenseIL,wieczorek2021unreasonable,flipreid} which almost saturated the person re-identification benchmarks. However, these results hinge on the crucial premise of \textit{supervised methodology}, where models undergo training on {em labelled} data pertaining to a specific domain. These approaches demand a substantial amount of annotated data, resulting in significant human effort and computational costs for data labelling. 
In addition, models tends to overfit to the training data and struggle to generalize to new environments or unseen individuals.

To tackle the domain adaptation issue associated with supervised person Re-ID, recent research has focused on \textit{unsupervised person re-identification}, which involves training Re-ID models using abundant unlabeled data. The performance of unsupervised person Re-ID tends to be generally inferior to that of supervised person Re-ID, due to the challenges posed by the absence of pairwise labelled data for training camera-invariant representations of individual features \cite{UDA_surv}. However, in recent years, this performance gap has been diminishing, showing promising results \cite{funs21, funs11}. 

With this premise, we have undertaken the task of conducting this survey with two primary objectives:

\begin{itemize}
  \item Aggregating the most significant publications in supervised person re-identification, categorizing them based on their techniques, to provide an overview of the state-of-the-art and highlighting the difficulties for further improvement along this direction.
  \item Synthesizing the latest developments from the past three years in unsupervised person re-identification, offering insights into the state-of-the-art advancements and shedding light on future trends in this emerging and promising area.
  
\end{itemize}

The structure of this survey unfolds as follows. In the next section (Section \ref{sec:research}) the methodology of the research and selection of the papers was presented.
In Section \ref{sec: sup_reid}, we delve into supervised approaches, categorizing them based on the loss function they employ. Following that, Section \ref{sec: Uns_reid} explores the latest three years of state-of-the-art works in unsupervised methods. Section \ref{sec:Experimental_result} presents the results of both approaches along with some key considerations. Finally, the paper concludes in Section \ref{sec:conclusions} with our findings and insights.

\section{Search and selection of papers}
\label{sec:research} 
The objective of the survey was to compare articles with an available open source implementation, so we started collecting works from 
Papers with Code \footnote{\url{https://paperswithcode.com/}}. 
Our specific query targeted ``Person Re-Identification'', pointing to the typical benchmarks used to track progress in the field, and comprising 
``Market1501'' \cite{Scalable_person_re}, ``DukeMTMC-reID'' \cite{Duke}, and ``MSMT17'' \cite{person_transfer}. 

We utilized the portal's capabilities to identify top-performing papers based on metrics such as mAP, Rank-1, Rank-5, and Rank-10.

Then, we manually examined each of these top papers, integrating the 
source references with previous works and background bibliography in order to obtain a complete and essentially self-contained vision of the domain,
and shape a cohesive narrative to unify our findings.

We first categorized
the works into supervised and unsupervised methods based on a well known and 
widely adopted classification. The articles dealing with supervised methods 
have been clustered in Feature Learning and Metric Learning approaches, according to the emphasis given by the authors to the different aspects of their methodology. Works in unsupervised learning, are divided into Unsupervised Domain Adaptation, aiming to transfer knowledge from a labelled source domain,
to an unlabelled target domain, and Fully Unsupervised learning,  learning to discriminate between individual images without any notion of prior semantic categories.

\section{Supervised person re-identification}\label{sec: sup_reid}
Supervised techniques play a crucial role in tackling the challenge of person re-identification, as they offer the capability to compute metrics for evaluating and assessing model performance.

\begin{figure*}[htb]
  \centering
  \includegraphics[width=0.7 \linewidth]{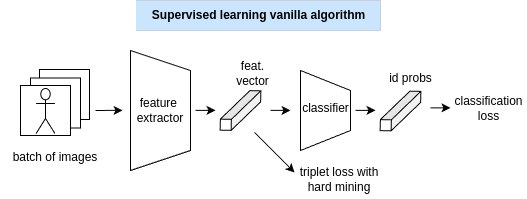}
  \caption{Vanilla model for supervised person ReID. During training Triplet loss works by selecting an anchor, a positive (same identity), and a negative (different identity) feature vector from the batch. For classification loss each identity in the training dataset is a class.}
  \label{fig:supervised_vanilla}
\end{figure*}

In the context of ReID, the main supervised strategy involves utilizing a labelled training set to train a deep neural network for building a robust representation able to identify individuals. Subsequently, the testing phase involves evaluating the model on previously unseen individuals present in two distinct datasets: the Gallery and Query sets.
The typical structure of a vanilla model for supervised person ReID is described In Figure \ref{fig:supervised_vanilla}. It consists of a backbone (e.g., CNN like ResNet) extracting visual features from images. Ideally, embeddings of the same individuals should have low distances between each other and high distances from others.
To this aim two tipical losses are applied: classification and triplet loss. The classification loss classifies the person ID based on the learned features; the triplet loss aims to minimize intra-class (same person) distances and maximize inter-class (different people) distances.

During the test stage, the network computes the representation of features for each image of the gallery and query sets. This leads to the creation of a distance matrix between every sample of both datasets. 


This section is divided into two parts: the first delves into the feature learning approach, exploring techniques to extract discriminative features from raw images, while the second focuses on the metric learning approach, looking for a similarity metric or distance function that can accurately quantify the similarity between any two feature vectors.
Both metric learning and feature learning play crucial roles but address different aspects of the problem. 

\subsection{Feature Learning}\label{sec:CL_DL}
Feature learning for ReID is a methodology centered on extracting distinctive and discriminative characteristics from raw images, aiming to effectively capture and represent key information relevant to the correct identification of individuals.

In a typical ReID feature learning model, input images are processed through a pre-trained CNN backbone, like ResNet50~\cite{he2016deep} or Inception V3~\cite{szegedy2016rethinking}, producing corresponding feature maps. These pre-trained models, trained on extensive datasets, enable the transfer of learned knowledge to new, potentially smaller, datasets for different tasks. Subsequently, specific techniques can be employed on the extracted feature maps to focus on so called {\em local} or {\em global features}, discussed hereafter.

Local features refer to fine-grained details and characteristics specific to different body parts, allowing for a more granular representation of individuals. Methods for extracting local features include partitioning either a single image or its features extracted from a CNN backbone network into multiple parts. These partitioning strategies pursue extracting on improving unique features in specific body parts and then bringing together these features from all parts to provide distinctive cues for each person's overall identity.

\begin{figure*}[htb]
  \centering
  \includegraphics[width=1.0 \linewidth]{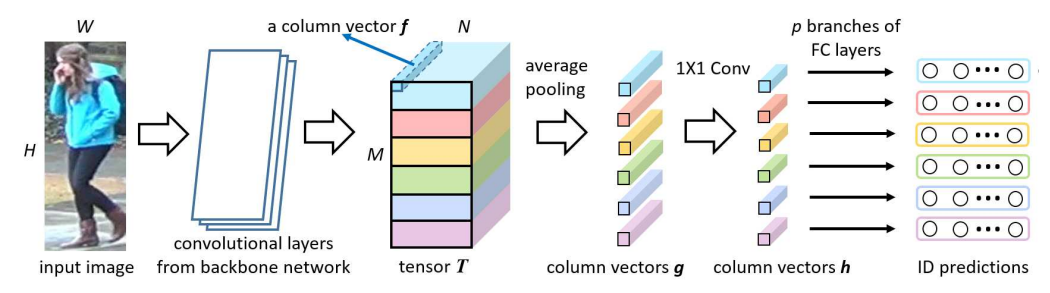}
  \caption{Structure of PCB. The feature maps extracted by the backbone network are segmented into six horizontal stripes, and each stripe undergoes horizontal average pooling, producing a single-channel vector for each stripe. Figure taken from \cite{sun2018beyond}}
  \label{PCB}
\end{figure*}

One of the first works that effectively utilizes local feature extraction is the Part-based Convolutional Baseline (PCB) \cite{sun2018beyond}, which is used as baseline in many other ReID works \cite{wang2018learning,sun2019local, chen2020salience,CorradoBCN23}. This method typically employs ResNet50 as the backbone network: to better learn part-level features, the resulting feature maps are then segmented into six horizontal stripes, and each stripe undergoes horizontal average pooling, producing a single-channel vector for each stripe, as shown in Figure \ref{PCB}. Subsequently, a 1x1 convolution is applied to each channel vector, and the reduced vectors are fed into separate classifiers. The primary objective of the training process is to minimize the sum of Cross-Entropy losses over the identification predictions of each classifier.

However, challenges such as occlusion and background clutter can hinder the accurate extraction of local features. Moreover, this process often fails to cover all discriminative information, since it almost exclusively focuses on specific parts with fixed semantics, 

Global features encompass the overall representation of an individual, providing a holistic view. While global features are advantageous for capturing the general appearance of a person, they may face limitations in addressing detailed variations and pose changes, especially in crowded environments.

In the realm of ReID, successful feature-learning strategies rely on the combined extraction of both local and global features. 
This integration is crucial because it leverages the strengths of each type of feature, allowing the model to benefit from the detailed discriminative power of local features while also capturing the general context and appearance through global features.

Typically, the extraction of global and local features occurs within a multi-branch network architecture, where one branch is commonly specialized in extracting global features and the other is dedicated to local features, as in the work by Wang et al.  \cite{wang2018learning}.

Wang et al. \cite{wang2018learning} highlight the limitation of consistent feature extraction at a uniform level of detail in previous works, such as the PCB baseline\cite{sun2018beyond}. They argue that this approach overlooks the opportunity to capture discriminative features across various levels of detail. To overcome this limitation, they propose a three-branch network architecture, where a unified backbone, ResNet50, generates feature maps. In this design, the global branch processes the feature maps through the last residual block of the backbone, introducing down-sampling via a stride-2 convolution layer, followed by global max-pooling, 1x1 convolution with batch normalization, and ReLU activation. Meanwhile, the two local branches leverage the original last residual block of the backbone. These branches uniformly partition the feature maps into multiple stripes, adjusting the number of parts across different branches to capture local features with diverse levels of detail. The outputs of these branches are split into 2 and 3 horizontal stripes, respectively, aiming to represent local features at distinct levels of detail.

Conversely, the work in \cite{class_2} improved the visual similarity probability of the PCB backbone \cite{PCB} by adding a second independent branch that processes spatio-temporal probability using the Histogram-Parzen (HP) method. This branch assesses the likelihood that images from two cameras taken at different times represent the same person, based on the time gap between captures. Additionally, to address rare events in walking trajectory and velocity of a person, a third module called joint metric, uses a mathematical logistic smoothing approach to adjust the probabilities and compute the combined likelihood that two images belong to the same ID given specific spatio-temporal information.
The techniques used for extracting global and local features may vary, adapting to the specific characteristics of the network. 
\cite{class_tripl_4} suggests a lightweight network that integrates global, part-based, and channel features within a cohesive multi-branch structure, leveraging the resource-efficient OSNet backbone \cite{OsNet}. Each branch's output undergoes batch normalization, followed by a fully connected layer corresponding to the number of classes. Embeddings extracted pre-batch normalization are optimized using a triplet loss, while those post-fully connected layer are optimized with a Cross-Entropy loss.
Recently, MSINet \cite{MSINet} was introduced, also relying on OsNet. This neural network employs multi-scale feature learning for re-identification tasks. The architecture consists of stacks of MSI submodules and downsampling blocks.
Each MSI submodule includes two separate branches of convolutional layers for handling different image resolutions with different receptive fields at a ratio of 3:1, allowing the model to process the same image through distinct pathways.
The results from these branches are then aggregated using an integration module before being 
fed into a downsampling block. The model demonstrated notable results in various supervised 
re-id tasks and performed well on person re-identification benchmarks.

In the subsequent subsections, we will illustrate some body parts dedicated techniques.

\subsubsection{Body parts learning}

Solutions that prioritize the extraction of semantic information related to various segments of a person in an image can significantly enhance the process of local feature learning. In this context, human semantic parsing proves to be particularly suitable.

Human semantic parsing is an image segmentation technique that partitions a person's image into specific elements like the head, torso, arms, and legs while preserving rich semantic detail \cite{yang2023deep}. Operating within the framework of image segmentation, this technique's ultimate goal is to classify each image pixel into distinct classes, with each class representing a specific human body part. Human semantic parsing enhances understanding by finely segmenting and categorizing various body parts, adopting CNNs for this purpose. CNNs for image segmentation and human semantic parsing can further integrate Atrous Spatial Pyramid Pooling (ASPP). ASPP employs dilated (atrous) convolutions at multiple rates, introducing gaps in the convolutional kernel to extend its receptive field. This technique empowers the network to gather multi-scale contextual information, fostering a more comprehensive understanding of objects at various scales within an input feature map \cite{chen2017rethinking}.

Significantly, the use of semantic segmentation in human parsing serves to amplify local visual cues in images, seamlessly aligning with the emphasis on local feature extraction for person identification tasks.

Leveraging human semantic parsing for local feature extraction, Kalayeh et al. \cite{kalayeh2018human} introduce a dual-branch architecture, utilizing one branch for local and the other for global feature extraction. In the global branch, the backbone retains the original Inception-V3, incorporating scale-up/scale-down operations on feature maps. Conversely, in the local branch they introduce a customized Inception-V3 tailored for human parsing, recognizing the crucial role of high-resolution final activations in human semantic parsing. To accomplish this, they reduce the stride of the grid reduction module in Inception-V3. Additionally, they replace the last global average pooling layer of Inception-V3 with Atrous Spatial Pyramid Pooling (ASPP) and a 1x1 convolution layer for classification. This modification yields probability maps for five body regions: foreground, head, upper body, lower body, and shoes. These maps undergo matrix multiplication with the global feature vector, and the resulting vectors, along with the global feature vector, are concatenated for identity inference using softmax cross-entropy loss in the classification head. 

Another body part technique is introduced by the authors of \cite{quispe2019improved}, which advocate for the utilization of saliency detection in ReID, a task centered on pinpointing visually interesting regions that immediately draw a human viewer's attention. Notably, Deep Learning Saliency methods, exemplified in \cite{li2016deepsaliency}, employ fully convolutional neural networks (FCNNs) to simultaneously capture the semantic attributes of salient objects through CNN. Specifically, \cite{li2016deepsaliency} utilizes a customized FCNN with shared convolutional layers for feature extraction and task-specific deconvolutional layers, optimizing them based on ground truth saliency/segmentation maps.

\begin{figure*}[htb]
  \centering
  \includegraphics[width=0.9 \linewidth]{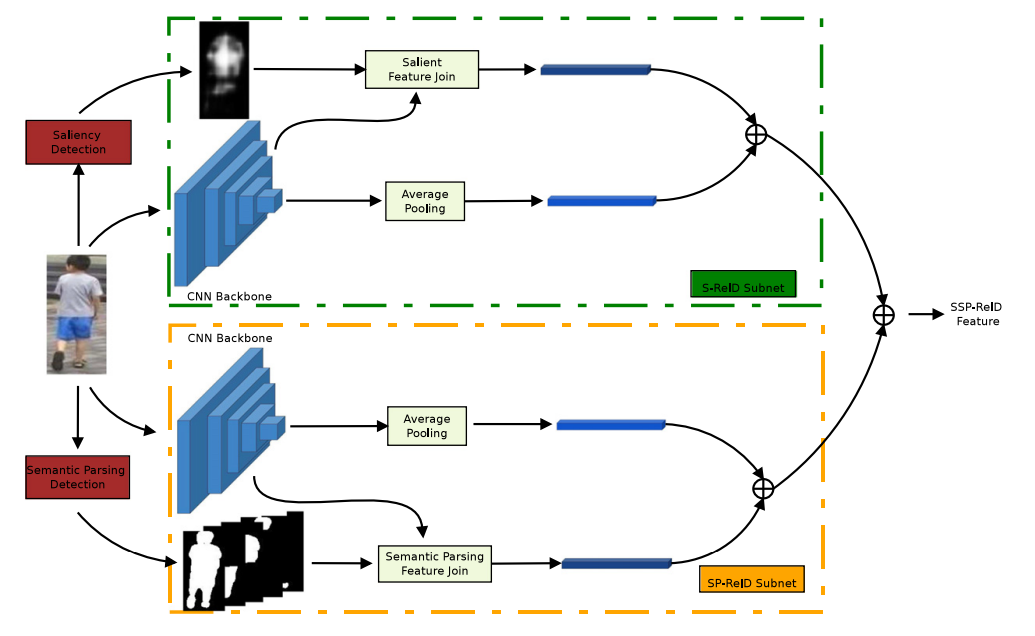}
  \caption{SSP-ReID framework. One branch is responsible for human semantic parsing, while the other for saliency detection. Figure taken from \cite{quispe2019improved}}
  \label{Saliency}
\end{figure*}

Acknowledging the potential of saliency detection to emphasize certain objects as discriminative cues in aiding the ReID process, Quispe et al. \cite{quispe2019improved} also acknowledge its limitation in dealing with occlusions, primarily due to its selective focus on specific areas within an image. To address this, they also integrate the local capabilities of human parsing in the Saliency-Semantic Parsing (SSP-ReID) framework \cite{quispe2019improved}, which leverages both human parsing and saliency detection (Figure \ref{Saliency}). With two streams sharing the same backbone architecture but without weight sharing, one of the subnetworks focuses on saliency features using \cite{li2016deepsaliency}. The other stream concentrates on semantic-parsing features through \cite{gong2017look}, extracting segmented maps for individual body parts. Each segmented map undergoes processing, computing the centre point for the respective body part. The evaluation process assesses centroid map quality using Euclidean metric and pixel-wise softmax loss, comparing each segmented image with its corresponding ground truth map. Both streams undergo mirrored preprocessing steps, combining saliency/semantic maps with intermediate feature maps through channel-wise element-wise product and average pooling. The result combines with the final backbone output, yielding a global saliency/semantic parsing map used for a final classification layer.

\subsubsection{Attention-based techniques}

Other feature-learning techniques leverage the application of visual attention to extract discriminative features. The fundamental objective of visual attention is to emulate the cognitive processes of the human visual system, zooming in on pivotal features within the input image \cite{hassanin2022visual}. Incorporated into deep learning models, attention mechanisms empower these models to concentrate selectively on particular regions or features within an image. The underlying concept involves the application of a learned attention mask to the feature maps. This not only aids in singling out a focused foreground but also amplifies distinct representations of the selected objects. Typically, traditional techniques adopt convolution operations on feature maps to generate attention masks.

Li et al. \cite{li2020scanet} propose the SCANet framework, which uses multi-scale convolutions to generate a foreground attention mask. The system consists of two parallel modules—Foreground Mask and Baseblock—both utilizing different kernel sizes to capture multi-scale information. The Foreground Mask module filters out background noise, and this process is repeated across layers. They also introduce a Semantic Consistency Loss to ensure consistent attention across scales, guiding the network to focus on foreground areas in shallow layers for better feature learning in deeper layers.

In their methodology, Zhu et al. \cite{zhu2020curriculum} strategically place the block responsible for attention mask computation within a two-branch architecture, aiming to enhance the features of both blocks with the same mask. The overall architecture includes a backbone ResNet50, a global branch, a local branch, and a Supervised Attention Module (SAM). Notably, SAM is intentionally inserted between the global and local branches to establish a correlation and bridge the connection between the two. In this configuration, the penultimate output of ResNet serves as the input for both the SAM module and the global branch. The SAM module creates an attention mask through a series of convolutional operations, refining the inputs for both global and local branches. The global branch processes its input through the application of the attention mask and further refines the result with the last layer of ResNet50 for ID classification. Simultaneously, the attention mask determines the input for the local branch by applying it to an intermediate result of the convolutional operations that produce the attention mask. Within the local branch, the previously mentioned PCB architecture \cite{sun2018beyond} is utilized. Noteworthy is the dual role of the convolutional layers employed in SAP, contributing not only to attention mask generation but also to ID inference. This dual functionality allows for the optimization of weights that generate the attention mask by emphasizing label-related discriminative information.

In an alternative approach, Sun et al. \cite{sun2019local} propose to obtain an attention mask adopting a non-local operator \cite{wang2018non} instead of convolutions. Unlike conventional convolutional layers that focus on local neighbourhoods, non-local operators collect features for each pixel, known as the local pixel, by taking into account the contribution of all other pixels in the image, termed non-local pixels. This technique enables the capture of long-range dependencies and relationships between spatially distant locations, a capability that traditional convolutions, limited to local neighbourhoods, do not possess.

In the LGMANet framework \cite{sun2019local}, ResNet50 serves as the backbone, feeding its output into two branches: the local-to-global and the multi-scale attention branch. The local-to-global branch addresses disparities in local and global features by subdividing downscaled feature maps into distinct horizontal stripes, reflecting the original backbone feature maps at different scales. On the other hand, the multi-scale attention branch generates attention feature maps using a non-local operation, calculating pixel features as weighted sums of features from all pixels in the input maps. The weighted sum involves exponentiation of the product of embedded local and non-local pixels, weighted by a linear embedding of the non-local pixel. The embeddings are generated through two different 1x1 convolutions for local and non-local pixels. The resulting attention mask is later applied to the backbone feature maps through an element-wise sum, producing the attention feature map. Finally, the outputs of both branches are pooled, summed, and sent to a fully connected layer for ID predictions.

\subsubsection{Feature-erasing techniques}

As highlighted in the study by \cite{chen2020salience}, although attention-based methods have advanced person re-identification, a significant challenge persists in effectively extracting discriminative salience features from diverse individuals. These methods primarily concentrate on capturing 'easy' features that contribute to reducing training loss over seen classes, rather than learning comprehensive details and concepts. This suggests that these approaches do not offer cues to enhance and diversify an individual's representation, posing a problem in ReID where the network needs to provide the richest and most diverse representations possible. To address this challenge, feature erasure methods have been proposed. These methods involve erasing the most discriminative part of feature maps to encourage the CNNs to detect intricate and potentially significant details in other image regions.

In their model, Benzine et al. \cite{benzine2021deep} incorporate a dedicated suppression procedure branch within a dual-branch architecture for identifying and eliminating the most discriminative features. This works introduces an Input-Erased branch (IE-branch) to the existing global and local branches. The IE-branch can be placed at any point between two convolutional layers in any branch. It takes the feature maps from the previous layer to compute a binary mask using channel-wise average pooling and min-max normalization, erasing pixels based on a threshold. The resulting mask is then applied to the feature maps through element-wise multiplication for each channel, resulting in the erased feature map.

Nonetheless, the authors in \cite{quispe2021top} highlight that the systematic erasure of features can introduce noise into the erased feature maps. This occurs due to false positives generated by removing regions that represent unique characteristics among different ID inputs.

To deal with this problem, they adopt in their framework \cite{quispe2021top} a regularizer branch that forces the features before the dropping step to be still discriminative for ReID. The overall architecture expands a ResNet50 backbone, with Global, Top DropBlock, and regularizer streams. The Top DropBlock stream uses two branches: one processes feature maps with convolutions, and the other generates a binary mask. This mask, formed from the backbone's feature maps by summing pixel absolute values across channels and row-wise averaging, eliminates less activated regions from the other branch through a dot product. Finally, the regularization stream, only used during training, appends a global average pooling layer after the feature maps computed before the application of the mask. This stream aims to safeguard against noise introduced by Top DropBlock, preserving relevant information and maintaining overall model performance. 

\subsubsection{Transformer-based techniques}
\begin{figure*}[htb]
  \centering
  \includegraphics[width=0.6 \linewidth]{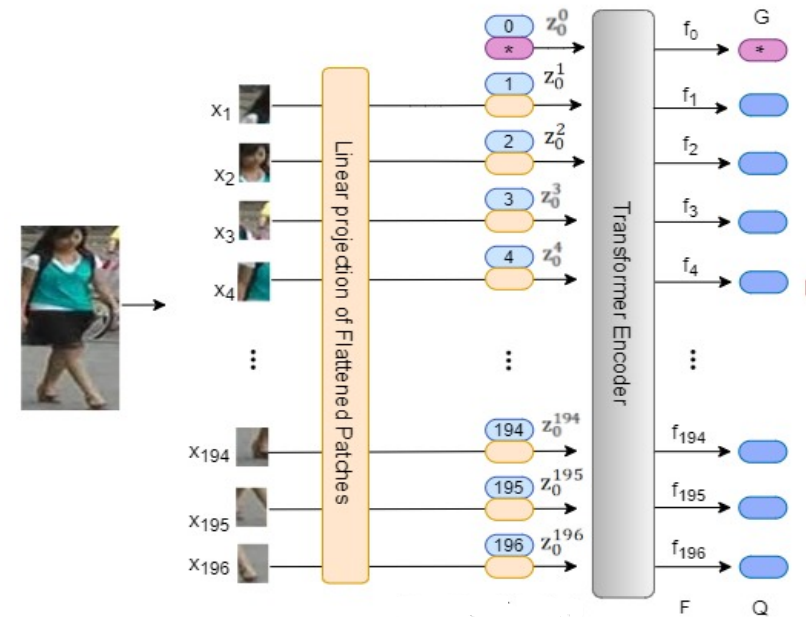}
  \caption{In the Vision Transformer (ViT) framework, the image is partitioned into a sequence of non-overlapping patches, forming the input for the ViT model. The model processes these patches, assigning each one a corresponding token. Additionally, a global token, denoted by the letter G and highlighted in violet, represents the overall information from the entire image within the sequence of tokens. Figure taken from \cite{sharma2021person}}
  \label{ViT}
\end{figure*}

While CNN-based methods have achieved remarkable success, they encounter limitations in processing only one local neighbourhood at a time. This restriction results in information loss on details due to convolution and downsampling operations, including pooling and strided convolution.  In contrast, Vision Transformers (ViT) \cite{dosovitskiy2020image} address these challenges by dividing images into non-overlapping patches. The sequence of patches serves as the input for the ViT, which employs multi-head self-attention within the ViT Block to capture inter-patch representations. The output of a ViT consists of a sequence of tokens, with each input patch assigned a token, alongside a global token, as shown in Figure \ref{ViT}. While each patch token encapsulates a representation that integrates both low-level and high-level visual features from its corresponding region in the input image, the global token aggregates information from the entire input image. This approach enables the ViT to capture holistic information about the overall scene, surpassing the limitations of local processing in CNNs and providing a more comprehensive understanding of the input data. 

In contrast to the previous ReID techniques where a backbone CNN leads to different branches for capturing local/global features, a ViT inherently encapsulates both types of information in the output token sequence. Given ViT's widespread success in various computer vision tasks \cite{wu2020visual}, \cite{wang2021pyramid}, \cite{chen2021pre} researchers have increasingly turned to ViT techniques for person re-identification as well, marking a departure from traditional approaches: \cite{he2021transreid} (Transreid),\cite{sharma2021person} (LA-Transformer), \cite{zhang2023completed}.

Visual Transformers for person re-identification have been explored for the first time in \cite{he2021transreid}. The work establishes a robust baseline framework centred on a pure transformer architecture. In this foundational approach, overlapped patches are employed, and solely the global embedding is utilized for ID inference.  To further enhance the robust feature learning, the authors introduce a module that shuffles the sequence order of local tokens and regroups them into distinct sets. Within each set, the global token is concatenated at the beginning. This random division ensures that each group does not contain local tokens from a specific part of the image, allowing each local feature within a group to cover patches from different body regions and so possess global discriminative capabilities. 

Sharma et al. \cite{sharma2021person} proposed a Transformer baseline with a PCB-inspired strategy. When a ViT encoder outputs  N+1 tokens, the [CLS] embedding is considered the global feature (G), and the remaining tokens are the local features (Q). Q and G are then combined using weighted averaging and arranged into a 2D spatial grid. Finally, a Locally Aware Network takes the row-wise average of the grid (or tensor), which is performed using average pooling and its output is then fed into a classification ensemble.

Zhang et al. \cite{zhang2023completed} also capitalize on the effectiveness of information from rigid stripe parts, akin to \cite{sun2018beyond}, but with an application to raw images. Their methodology involves segmenting the input image into multiple stripes, which are then processed by the Completed Part Transformer (CPT) consisting of intra-part and part-global layers. These two layers are specifically crafted to take into account long-range dependencies from the aspects of the intra-part interaction and the part-global interaction. The intra-part layer facilitates interaction among patch tokens within each stripe, employing a self-attention mechanism to capture long-range dependencies. Simultaneously, the global token engages with all patch tokens across the image in the part-global layer. Each stripe, represented by a feature vector, utilizes a cross-attention mechanism to aggregate global information from all patch tokens in the pedestrian image.  

\subsubsection{Cross-Domain adaptation}
Cross-domain adaptation refers to the process of adapting a model trained on a source domain to perform well on a different but related target domain. This approach typically assumes some level of supervision in the target domain, which can vary from having a few labeled examples (semi-supervised) to having style or attribute annotations (weakly supervised). We shall discuss fully 
Unsupervised Domain Adaptation in Section \ref{sec:UDA}.

Cross-domain adaptation methods often involve:
\begin{itemize}
\item Transfer Learning: Leveraging knowledge from the source domain to improve performance in the target domain. This might involve fine-tuning a pre-trained model on a small set of labeled data from the target domain \cite{Deep_transfer_2018,person_transfer}.
\item Domain Mapping: Transforming data from the source domain to appear as if it came from the target domain (or vice versa), thereby reducing the domain gap. Generative models, like GANs, are commonly used for this purpose \cite{GAN2,GAN3,GAN4}.
\item Feature Alignment: Modifying the feature space to minimize the discrepancy between source and target domain distributions in the feature space. This process involves adjusting the features extracted from both domains so they become more similar, making it easier for a machine learning model to perform accurately on the target domain without requiring extensive labeled data from that domain \cite{Deep_CORAL, Domain_adaptation_survey_2018}.
\end{itemize}
Combining cross-domain adaptation with data augmentation offers a synergistic approach to handling domain shifts. Data augmentation can increase the diversity and volume of training data, simulating conditions that may not have been covered by the original dataset. This enriched dataset can then serve as a better foundation for cross-domain adaptation techniques, which fine-tune the model to align more closely with the target domain's characteristics.

For example, generative models like GANs or Diffusion Models can be used not only for feature alignment but also for generating synthetic images that bridge the gap between the source and target domains. These images can then be used as part of the training set, effectively augmenting the data with examples that are directly relevant to the target domain's distribution.

A different perspective on the use of generative techniques for Person-ReID has been recently advocated in \cite{sensor_generative},
where the latent representation of different individuals is learned as the shared information necessary to condition the generation of images of the given person from different views, and in different contexts. This approach enables the separation of an individual's identity from other instance-specific information, such as pose and background.

\subsubsection{Final considerations and future directions}
The exploration of feature learning methods for person re-identification highlights the importance of combining local features, which capture fine details, with global features that provide contextual understanding. This integration is key to addressing diverse scenarios, though challenges in achieving effective re-identification remain.

All techniques that have been discussed have advantages and disadvantages.

Multi-branch networks prove effective for diverse feature extraction, yet the trade-off between feature richness and computational demands requires careful consideration, impacting real-time efficiency. 

Body-parts learning usually improves feature discrimination and is robust to occlusions; however, localized feature extraction requires accurate part detection and, especially, it is inconsistent across different camera angles.

Erasure-methods typically enhance generalization and reduce the risk of overfitting, but they also introduce a high risk of information loss, making the technique largely dependent on the actual erasing strategy. 

The attention-mechanism, especially in conjunction with visual transformers, allow the model to learn in and end-to-end way to 
selective focus on important features. This typically results in ReID models that are more adaptive, robust to background clutter, and able to capture the global context. On the downside, these models are usually data-hungry, and prone to overfitting on small datasets.

Generalization across domains still remains the major issue in person 
Person ReID; domain adaptation techniques directly address this topic,
and usually allow a model to increase transferability across different 
camera networks or environments, reducing the dependency on labeled data.

Despite improvements, there can still be a significant performance drop when 
moving from one domain to another, especially if the differences between 
the domains are substantial (e.g., different lighting or camera angles).

In conclusion, the recent trend in person re-identification research leans 
towards transformer-based methods. These models, by inherently integrating 
global and local feature extraction within their architecture, 
are effectively suitable for feature learning in ReID, 
addressing long-range dependencies better than CNN, 
despite the associated challenges in training intricacies and resource demands.

\subsection{Metric Learning Approach}\label{sec:sup_reid}
As we previously stated in the introduction, Person Re-identification is an image-based instance retrieval problem that essentially extends the classification problem from closed-world settings to open-world scenarios. Metric learning provides key techniques to navigate these challenges, making it the most actively researched area in the Person Re-identification field today.

The main idea of metric learning is to employ a specific loss function during training to guide the feature extractor towards favoring a clustered structure of the embedding. The state of the art in Re-ID losses involves the combined use of three criterion functions: Identity, Verification, and Triplet losses.

\subsubsection{ Classification strategy}
In typical classification problems, cross-entropy loss is highly effective in facilitating the learning of high-level and semantically rich embedding representations within the network to distinguish instances of different classes (\cite{Horiguchi_2019}). 
The work presented on \cite{Zhai_2019_CVPR_Workshops}  serves as an example of using cross-entropy for identity separations as a classification strategy. In this context, each unique ID represents an individual class, and the network's output is a vector as long as the number of identities in the training set. This loss is known in the literature as identity loss:

\begin{equation}
    L(p, q) = -\sum_{i=1}^{C} p_i \cdot \log(q_i)
\end{equation}
where:
\begin{itemize}
    \item $p_i$ is the true probability of the input belonging to class $i$.
    \item $q_i$ is the predicted probability of the input belonging to class $i$.
    \item $C$ is the number of identities .
\end{itemize}

The network learns from training data to classify every instance. Then, during inference, the last layer is removed and the outputs of the second-to-last layer is used to compute a robust matrix of distances between query and gallery samples of the new people in the testing sets. 
The shifts among training and testing tasks is a discrepancy that need to be addressed.

\subsubsection{Verification Strategy}
A practical solution to generalize the embedding of an instance is to apply a second training stage to the model by solving the corresponding binary classification problem: {\em Person verification}. A pair of images is input into the same model, and the outputted similarity score of the embeddings indicates whether the two images correspond to the same individual. The most common loss function used is the verification loss, which is typically expressed in the form of binary cross-entropy (BCE):

\begin{equation}
    L_{\text{BCE}} = -\frac{1}{N} \sum_{i=1}^{N} \left[ y_i \cdot \log(p_i) + (1 \!-\! y_i) \cdot \log(1\! -\! p_i) \right]
\end{equation}
where:
\begin{itemize}
    \item $N$ is the number of sample pairs.
    \item $y_i$ is the binary label indicating whether the pair is of the same person (1 if yes, 0 if no).
    \item $p_i$ represents the predicted probability of the pair being of the same person.
\end{itemize}

It is also common applying to the model a single training stage where the verification loss is merged with the classification loss to enhance a model's performance and to leverage their respective advantages: while the classification loss creates separation among distinct classes, the verification loss minimizes the distance within the same class; did so \cite{deng2018image} and \cite{chen2018group}. 

\subsubsection{Triplet Strategy}

Contrastive learning has taken place into metric learning and ReId problem as well. 
The general idea is to minimize the distance between the embeddings of the same person ( the intra-class distance ) and maximize the distance between different embeddings identities (the inter-class distance).
In the family of contrastive losses, Triplet losses is the main framework. In recent works it replaced the verification loss and it is often used in combination with Identity losses. In \cite{hermans2017defense} is showed how the triplet loss can enforcing directly a clustered embedding structure.
Given an embedding $z_i = f_{\theta}(x_i)$ of a given image $x_i$ trough the neural network $f_{\theta}$. 

A triplet is a tuple that contains one anchor sample $z_a$, one positive sample $z_p$ and one negative sample $z_n$ from a different identity. The general loss formulation is
\begin{equation}\label{eq:triplet_loss}
L_R(z_a,z_p,z_n) = max(0, m + d(z_q,z_p)  - d(z_q,z_n)) 
\end{equation}

Since the input images are normalized to fall within the $[0,1]$ range, the output resides on a hypersphere, and $d(\cdot)$ measures the Euclidean distance between two samples. The parameter $m$ represents the margin, ensuring that negative pairs are at least this margin value distant from positive pairs. Moreover, the $max(0, \cdot)$, which essentially denotes the Hinge Loss, can be substituted with $\ln(1 + \exp(\cdot))$, a smooth approximation of the max function that ensures training stability. This function is known in the literature as the log-sum-exp loss.

The study conducted by \cite{hermans2017defense} introduced a significant advancement in triplet loss methodology. Specifically, it involves the selection of the most challenging positive and negative samples from each batch, referred to as $z_a$, when constructing triplets for loss computation:

\begin{equation}\label{eq:triplet_hard_loss}
\begin{split}
L_R(z_a,z_p,z_n) = & max(0, m + \max_{p=1...k}( d(z_q,z_p) ) \\
                  & - \min_{n=1...k} d(z_q,z_n)) 
\end{split}
\end{equation}

Over the years, research on the triplet loss has been conducted to improve intra-class and inter-class distances of the overall output. The work
\cite{zhang2018person} tries to weight more negatives samples to positives using a focal loss applied to the triplet criterion. 
Another way to get similar results is to calculate the centroid of a class of embeddings and adjust the anchor towards it. Moving samples closer to the center of their class and farther from the center of negative classes makes the representation more cluster-consistent and makes the model more robust against outliers.
One of the early attempts in this direction was the work by Si et al. \cite{si2019compact} (2019), which investigated an identity+triplet loss with two regularizers based on centroid differences for inter and intra-class distances, respectively.

Wieczorek et al. ~\cite{wieczorek2021unreasonable} compared the distances between anchors and both positive and negative centroids. They used ResNet50 as a feature extractor, applied global average pooling to create 2048D embeddings, and combined three losses: triplet loss on raw embeddings, center loss as an auxiliary loss, and classification loss on batch-normalized embeddings.

A similar approach was also employed in~\cite{zhao2022multi}, where the most challenging positive and negative anchor-centroid distances within a mini-batch were identified. More recently, \cite{alnissany2023modified} divided the triplet loss into two distinct differences, squared the distances to amplify them, and then multiplied both values by two hyper-parameters $\omega_1$ and $\omega_2$ such that $\omega_1 = ( 1- \omega_2)$; choosing $\omega_1 < \omega_2$ has the effect of more heavily penalizing inter-class distance. 
Another direction was proposed by \cite{chen2017beyond} which use a quadruplet loss made by two negatives samples: $(a,p,n_1,n_2)$. 
Zhang et al.~\cite{zhang2021triplet} introduced an N-tuple loss to enable the joint optimization of multiple instances.

The efficiency of sampling when generating effective triplets is a critical aspect of learning with such losses, given that the number of potential triplets increases cubically with the number of images. Sampling without selecting optimal combinations can render the model training less effective, thereby diminishing its generalization capabilities. For this reason, researchers have developed various methods for efficient triplet selection, aiming to identify informative triplets. This approach is supported by multiple studies, such as~\cite{hermans2017defense,pos_emb,mishchuk2017working}.

Specifically, \cite{pos_emb} demonstrates, in the context of re-identification, the importance of sampling good moderate triplets that successfully adhere to the equations governing triplet distances:

\begin{equation}\label{eq:triplet_loss_constraints}
d(z_q,z_p) < d(z_q,z_n) < d(z_q,z_p) + m
\end{equation}

Hermans et al. \cite{hermans2017defense} proposed randomly sampling a batch composed of $K$ images from $P$ distinct individuals. For
each sample in the batch, the approach involves finding the hardest positive and the hardest negative samples {\em within the batch} when
forming the triplets for computing the loss.

Another strategy was presented by \cite{mishchuk2017working}. This method involves generating a batch with positive pairs in the form $(x_a, x_p)$, followed by calculating a distance matrix from which the closest non-matching descriptor is selected for each pair. Finally, among two negative candidates, the hardest one is chosen.

\subsubsection{Other loss strategies}
Many authors employ ad-hoc losses or combine them with additional features to improve performance. For example, in \cite{class_tripl_5}, a multi-branch network for scene disentangling is proposed, consisting of a main branch and two sub-networks to handle occlusion and scale variations. During training, the main branch processes the original image, while one sub-network applies random erasing and the other applies random zoom. All branches are trained using classification and triplet losses, with KL divergence aligning their outputs. During testing, images are randomly augmented and processed through all branches, and the resulting feature vectors are concatenated for the final re-identification task.

Similarly, \cite{flipreid} addresses the inconsistency issue between training and inference with an additional loss that reduces the gap between the mean feature vectors of multiple augmented samples. In their approach, triplet and identity losses are combined with a mean squared error loss between the feature vectors of original and flipped images during training. When image pairs (original and flipped) are fed into the FlipReID model, the flipping loss minimizes the mean squared error between their feature vectors. As a result, the model performs consistently whether test-time augmentation is used or not.

In another approach, \cite{contr_1} introduces a viewpoint-invariant ad-hoc loss. Here, the backbone generates a feature vector, which is passed to a classifier and used in triplet losses. The classifier applies cross-entropy to ID labels and classifies feature vectors into negative (high viewpoint similarity, different IDs) or positive (low viewpoint similarity, same ID) pairs. A variant of the L1 distance measures viewpoint similarity, and vectors are stored for each person. To enhance viewpoint robustness, a viewpoint discriminator module processes the feature vector, projects it into a lower space, and is trained using a contrastive temperature-scaled loss. A gradient reverse layer helps the feature extractor ignore redundant viewpoint information, while a hard-mined triplet contrastive loss based on memory further refines the feature extractor.

\subsubsection{CLIP-based person ReId}


A recent and promising approach involves leveraging advanced image-to-text embedding techniques to group images based on their visual similarities. A notable tool in this domain is OpenAI's Contrastive Language–Image Pre-training (CLIP) \cite{clip_main}. CLIP can be adapted to both supervised and unsupervised settings, depending on its application.

In a supervised context, CLIP is fine-tuned on a labeled dataset to master the specific task of matching different images of the same individual, making use of its pre-trained capabilities on a vast corpus of image-text pairs (i.e. \cite{yan2023clip,re-ID_CLIP,li2023clipreid} ). 

For instance, in \cite{li2023clipreid}, a two-stage training approach was used. In the first stage, embeddings for each person were generated using both text and image encoders, with the encoders kept frozen as part of the loss function. This allowed the text encoder to accurately describe each person. In the second stage, these embeddings were combined with a textual prompt and used with the frozen text encoder to improve the image encoder's ability to perform fine-grained person re-identification.

CLIP has also been used
in the context of video based Person ReId \cite{TF-CLIP}; specifically, the authors apply a
a pre-trained CLIP visual encoder to all video sequences
of a given identity to extract sequence features, and then 
average them to obtain an identity-level representation that they call CLIP-Memory. This approach allows to replace the text encoder
enablin Text-Free CLIP-based person ReID.

The CLIP technology can also be employed in an unsupervised framework, without directly relying on labeled data for person identification. Thanks to CLIP's capacity to understand and correlate images with textual descriptions, it can be used to generate pseudo labels through clustering of CLIP embeddings. These pseudo labels are then utilized to train a model for person ReID (i.e. in \cite{shaoZDW023}).

Moreover, CLIP's flexible nature facilitates semi-supervised or hybrid methodologies, where it is initially fine-tuned on a smaller set of labeled data and then applied to a larger set of unlabeled data. This process enhances its performance through self-learning or pseudo-labeling techniques. 

In conclusion, the application of CLIP for person ReID is versatile and can be customized to accommodate supervised, unsupervised, or hybrid learning paradigms, depending on the dataset available and the specific needs of the task.

\section{Unsupervised person re-identification}\label{sec: Uns_reid}
The unsupervised techniques for person re-identification can be broadly categorised into two main tasks: unsupervised domain adaptation (UDA) and unsupervised image encoder learning also known as fully unsupervised learning.
UDA typically requires labeled data in the source domain but does not require labeled data in the target domain; fully unsupervised learning aims to learn useful patterns or representations directly from unlabeled data without any domain adaptation. We categorize such approaches below.

\subsection{Unsupervised Domain Adaptation methods}\label{sec:UDA}
Unsupervised Domain Adaptation (UDA) techniques encompass the process of initially pre-training a model with labelled data from a source domain, followed by fine-tuning on an unlabeled target domain. The objective is to harness the knowledge acquired from the labelled source domain, thereby enhancing the model's performance when confronted with an unlabeled target domain.

Following this pre-training phase, the pre-trained backbone is utilized to extract features for unlabeled samples from the target domain. To introduce a form of supervision for these unlabeled samples, a clustering algorithm, such as K-means \cite{kanungo2002efficient} or DBSCAN \cite{ester1996density}, is applied to the extracted feature vectors. This clustering step serves as a means to generate pseudo-labels for each sample in the unlabeled set.

The term ``pseudo-labels'' is employed to emphasize that these generated labels are not ground truth annotations but rather labels assigned through a clustering process. These pseudo-labels are then used to guide the subsequent training of the feature extractor. For instance, they can be employed to calculate metric or contrastive loss, which necessitate labelled data to supervise the learning process.

The UDA clustering methodology alternates between generating pseudo classes through clustering target-domain instances and training the network using the generated pseudo-classes. This iterative process allows the source-domain pre-trained network to adapt effectively, capturing the inter-sample relations in the target domain based on the generated pseudo-labels \cite{fan2018unsupervised}.

Nevertheless, the primary challenge associated with UDA clustering techniques lies in the quality of the generated pseudo-labels, leading to predictions tainted by noise. This noise predominantly stems from two distinct sources. 

Firstly, the domain gap between the source and target domains, marked by a lack of shared identities and variations in viewpoint, light conditions, and background clutter \cite{lin2021unsupervised}.

Secondly, the inherent characteristics of clustering algorithms contribute to the challenge as they assign coarse-grained labels, grouping samples from the same cluster under the same identity \cite{ wang2022refining}. These clustering algorithms produce hard labels, which correspond to labels with 100\% confidence. This lack of flexibility makes the model less robust to variations in the target domain, amplifying the risk of misclassification errors that may permeate the entire training process.

Mitigating the impact of noisy pseudo-labels has emerged as a pressing concern within the realm of UDA ReID \cite{zhao2022exploiting}. A common solution is to introduce soft pseudo labels to better guide the training process. Soft pseudo labels differ from their hard counterparts in that they represent the probability distribution of an instance belonging to various classes rather than a hard classification. Calculating soft pseudo labels can be achieved through a fully connected layer. This is particularly viable because clustering algorithms provide the number of clusters, equivalent to the number of classes. Leveraging this information, a fully connected layer, equipped with nodes corresponding to the number of pseudo-classes, is applied to produce probabilities indicating the likelihood of each class for the input feature maps. The probabilities generated act as soft pseudo-labels and can be refined by utilizing one-hot encoding derived from the hard pseudo-labels assigned through the clustering process. Incorporating cross-entropy for refinement treats the pseudo-labels as the ground truth, compelling the soft pseudo-label probabilities to closely align with their corresponding hard pseudo-labels.

The flexibility of soft pseudo-labels allows for a more nuanced representation of uncertainty and a more robust adaptation in the face of domain shifts, especially crucial in the face of noisy pseudo-labelling in UDA scenarios.

In recent years, a variety of strategies have surfaced to address the significant challenge posed by noisy pseudo-labels. This issue is pivotal for devising effective solutions, particularly in the context of UDA approaches for ReID. These strategies strive to improve the quality of pseudo-labels by mitigating the influence of noise. In the subsequent sections of this chapter, we will delve into diverse techniques employed to achieve this objective.

\subsubsection{Mean Teacher-Student}
In the domain of unsupervised learning, the mean teacher-student framework represents a learning paradigm comprising two networks: the student and the teacher networks. Although these models share the same initial configuration, the optimization of weights follows distinct approaches for each network. Usually,
the student network is updated through traditional backpropagation techniques, while the teacher network evolves as a ``smoother'' version of the student model,
updating its weights according to an exponential moving average (EMA) of the latter. 
The ultimate objective is to jointly train both models mutually enhancing each other.

In the realm of UDA for ReID, the mean teacher-student framework stands out as a widely embraced backbone feature extractor \cite{soft_labels_UDA}, \cite{guo2022jac}, \cite{cheng2022hybrid}, \cite{ding2022learning}, \cite{wang2022learning} due to its effectiveness in leveraging the teacher network's reliability to improve the generation of pseudo-labels, facilitating more robust feature extraction.

One of the pioneering studies to introduce this paradigm is the work by Tarvainen et al. \cite{tarvainen2017mean}, initially applied in the context of generic semi-supervised learning. However, the authors themselves observed that it can be readily adapted to unsupervised scenarios.

\begin{figure*}[htb]
  \centering
  \includegraphics[width=0.8 \linewidth]{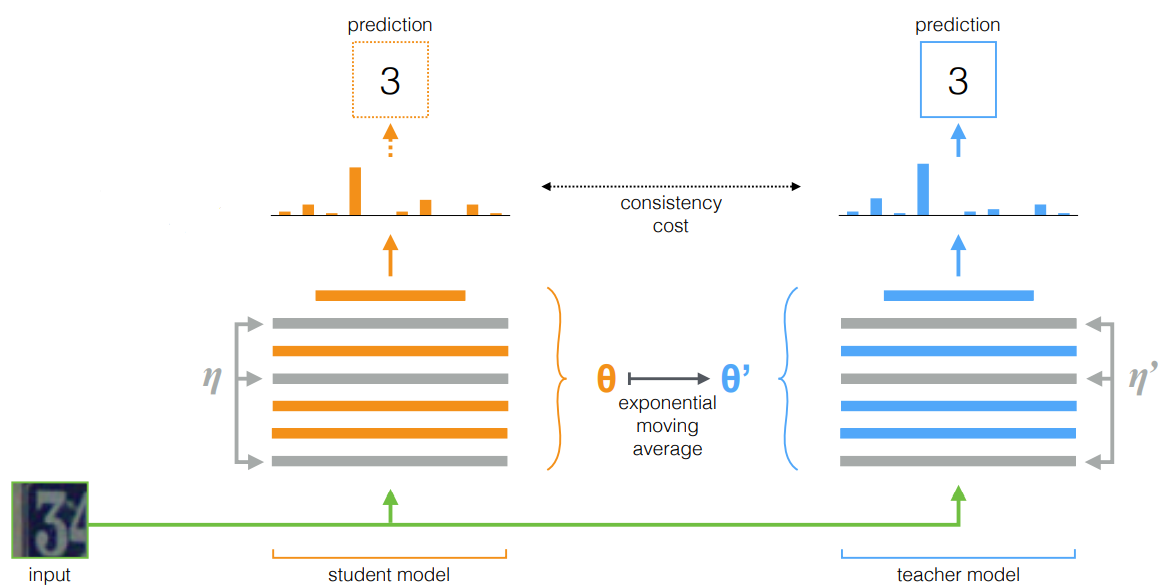}
  \caption{In the mean teacher-student framework, both models output the prediction for the same input (corrupted with different noise $\eta$ and $\eta'$). Then the consistency loss is calculated, to align the two predictions. Finally, the student model is updated via gradient descent, while the teacher model via exponential moving average (EMA). Figure taken from \cite{tarvainen2017mean}}
  \label{EMA}
\end{figure*}

In this framework, the student and the teacher networks start with identical configurations and receive the same input with different noise applications, ensuring a common input corrupted by diverse noise for both models. Subsequently, the predictions of both models undergo a comparison through a consistency loss, designed to minimize disparities between these predictions. Finally, the model parameters are refreshed: the student weights undergo optimization via gradient descent, while the teacher weights are updated through a weighted sum. This summation includes the teacher's parameters from the preceding training iteration and the parameters of the student model from the current training iteration, with a moving average approach (see Figure \ref{EMA}).

Here, a training iteration refers to a single gradient update during an epoch. Representing the parameters of the teacher network as $\theta'$ and the parameters of the student network as $\theta$, the optimization process for the teacher network weights adheres to the following equation:

\begin{equation}\label{eq:EMA}
\theta'_t = \alpha \theta'_{t-1} + (1 - \alpha) \theta_t
\end{equation}

where $\alpha$ is a smoothing coefficient hyperparameter and $t$ indicates the training iteration. With EMA, rather than directly sharing weights with the student model, the teacher network leverages its accumulated weights from prior iterations and combines them with the weights of the student model, which are optimized through gradient descent. This method aggregates information after each step, enhancing model accuracy compared to using final weights directly, as proposed by \cite{polyak1992acceleration}.  Upon completion of training, both model outputs can be employed for predictions; however, the teacher's prediction is generally more reliable, since it comprises an average between the optimized model of the student and its previous parameters.

An interesting variant is the Mutual Mean Teaching (MMT) framework 
 \cite{soft_labels_UDA}.
MMT is composed of two mean teacher-student networks that incorporate a modified version of the traditional triplet loss to refine pseudo-labels more effectively. In this approach, the two student networks generate hard pseudo labels, while the two teacher networks produce soft pseudo labels. This alternating strategy is implemented to prevent collaborative bias between the networks. The teacher network is then updated using a soft cross-entropy, where the soft predictions of one network are compared with those of the other. Additionally, the authors utilize a variation of the traditional triplet loss, comparing the softmax triplets from both networks through cross-entropy as follows: 

\begin{equation}
\text{Triplet}_{\text{soft}}(i,j) = -\sum_i T_j \log(T_i)
\end{equation}

The term $\text{Triplet}_\text{soft}(i,j)$ represents the soft triplet loss for the triplets of the $i$ network with the ground truth being the triplets of the $j$ network. $T_i$ and $T_j$ denote the soft triplets of the two networks, calculated as:

\begin{equation}
\text{$T_i$} = \frac{e^{||x_i - x_i^-||}}{e^{||x_i - x_i^+||} + e^{||x_i - x_i^-||}}
\end{equation}

Here, $x_i$ is the anchor feature vector computed by the $i$ network, and $x_i^-$ and $x_i^+$ are the negative and positive feature vectors, respectively, computed by the $i$ network. The resulting softmax $T_i$ value serves as a soft triplet label, approaching 1 as the model effectively separates negative pairs. In this study, $i$ and $j$ correspond to the two mean teacher-student networks, and for each, the soft triplet loss is computed with the ground truth being the softmax triplets calculated by the other network.


Guo et al. \cite{guo2022jac} also introduce a framework featuring two mean teacher-student networks, namely the Joint Learning with Adaptive Exploration and Concise Attention Network (JAC-Net). This framework additionally adopts a joint learning network, whose parameters are derived from averaging the teacher models' parameters of both networks. This joint network is introduced to exploit the knowledge of both models to prevent noise pseudo-label expansion during the training of the overall framework. JAC-Net encompasses two identical mean teacher-student networks, generating both hard and soft pseudo-labels through the K-means algorithm. These networks are guided by the soft predictions of the joint learning network, 

In the JAC-Net framework, the soft predictions of each network are compared with those of the joint learning model using cross-entropy and triplet loss, similar to \cite{soft_labels_UDA}. Notably, for the triplet loss, cross-entropy is computed based on the sum of distances between the anchor and the respective hardest positive and negative samples, omitting the softmax operation.

The overall loss function consists of the combined triplet and classification losses for both networks, with their contributions weighted by their discriminability. This comprehensive loss function integrates both triplet and classification losses across the networks, adjusting their impact based on discriminability. This adjustment accounts for the varying influences attributed to the quality of the generated clusters. Discriminability is evaluated using a coefficient that includes two dispersion indices for each network: one for intra-class dispersion, measured by the distance of feature samples from their average within clusters, and another for inter-cluster dispersion, gauged by the differences between the average features of a class and each image feature within that class.

\subsubsection{Memory-based contrastive techniques}
Memory-based contrastive techniques utilize memory banks to retain feature vectors computed during training, preserving the entirety of training data feature vectors. This enables the application of a clustering algorithm on the complete dataset, distinguishing it from non-memory approaches that solely cluster samples within each batch.

In a standard framework, a clustering algorithm is utilized to cluster all image features in the memory bank, generating pseudo labels. Subsequently, the entire model is trained using a contrastive loss, which is usually the softmax form of the InfoNCE loss \cite{oord2018representation}. For a given set of \(N\) samples, the softmax InfoNCE loss of a sample \(x_i\) belonging to the set is computed as:

\begin{equation}\label{eq:infoNCE}
L_{\text{InfoNCE}} = -\log \left( \frac{\exp(\text{sim}(x_i, x_j) / \tau)}{\sum_{k=1}^{N}\exp(\text{sim}(x_i, x_k) / \tau)} \right)
\end{equation}

where \( x_j \) is a positive sample from the set, \( \text{sim}(x_i, x_k) \) is a similarity score (e.g., cosine similarity) between the pair \( (x_i, x_j) \), and \( \tau \) is a temperature parameter.

This contrastive loss encourages a feature extractor to learn meaningful representations by maximizing the similarity between positive pairs (instances from the same category) and minimizing the similarity with other instances in the dataset. In other words, it guides the model to pull together features of similar instances while pushing apart features of dissimilar instances.

In memory-based contrastive methods, the utilization of a memory bank is crucial for mining hard positive and negative examples across all the training set, thereby enhancing the contrastive power of the model with a greater variety of samples and thus strengthening the pseudo-labels' reliability. In contrast, without a memory bank, the contrastive loss can only be computed using samples within the batch, limiting its effectiveness in capturing diverse instances and potentially introducing noise into the pseudo labels, which could hinder the training efficacy of the deep model.

Memory-based contrastive techniques exhibit variations primarily in their approach to managing memory. This distinction pertains to the selection of feature vectors that are stored in memory, consequently influencing which samples are considered in the computation of the contrastive loss. 

Ge et al. \cite{ge2020self} highlight the importance of storing in memory also the outliers obtained by the clustering algorithms. Clustering techniques like DBSCAN categorize training data into well-defined clusters and un-clustered outlier instances. These outliers are often discarded during training to uphold the reliability of the generated pseudo labels. However, un-clustered outliers often contain challenging yet valuable training examples for person Re-Id tasks, and simply abandoning them may significantly impair recognition performance.
In addition, the authors also point out the importance of using source domain samples during the training, which they consider very valuable due to the presence of ground truth labels. This inclusion is seen as instrumental in optimizing the feature extraction process.

For these reasons, they propose a framework \cite{ge2020self} with a memory that stores source domain class centroids, target-domain cluster centroids, and features of target-domain clustered and un-clustered instances.

At the start of each training epoch, the DBSCAN algorithm is applied to the unlabeled clustered and unclustered feature vectors in memory, to subsequently update the pseudo-labels of target domain feature vectors stored in memory, computing also the centroids for each cluster as the mean of the feature vectors belonging to it. Subsequently, a backbone CNN calculates feature vectors for each batch of data, encompassing samples from both the source and target domains. The source domain samples possess ground truth labels, while the target domain samples are equipped with pseudo-labels. Both types of labels enable the computation of the centroid for each class, determined as the mean of the feature vectors associated with that class. Then, for each sample in a batch is computed the contrastive loss: the positive sample is selected as the supervised centroid, unsupervised centroid, or as the corresponding outlier in memory on the sample being supervised, unsupervised, or an outlier feature vector respectively. Finally, the centroids for samples of both domains and the feature vectors of the target domain are updated in memory.


Song et al. \cite{song2022dual} critique works such as \cite{ge2020self} for solely focusing on aligning instance features with their respective cluster centroids. According to them, this approach neglects the complex relationships among instances within the same class, an important aspect considering the significant intra-class semantic differences found in person ReID datasets. These intra-class differences include a variety of visual aspects associated with the same identity or class, such as variations in clothing, pose, lighting, and background, which are critical for attaining high accuracy in ReID.

In response, they introduce the Dual Prototype Contrastive learning strategy, which adopts a contrastive loss that explores intra-cluster relationships. Their framework utilizes two memory banks for target domain images to compute two contrastive losses. The first memory bank stores cluster centroids computed at the beginning of each epoch through the DBSCAN algorithm, employed by the first contrastive loss as in \cite{ge2020self}. The second memory bank saves hard positive prototypes of each cluster, utilized in the second contrastive loss to reinforce relationships between inter-instances within the same class. 

The hard positive prototypes are initialized in memory at each epoch's start by randomly selecting one feature from each pseudo cluster after the clustering step. Subsequently, hard-positive prototypes are updated by the query feature sharing the same pseudo label in the batch while having the lowest similarity with their corresponding prototype in the hard-positive prototype memory bank. This contrastive loss ensures each query in the mini-batch is close to its corresponding hard-positive prototype while being distant from other prototypes, enhancing intra-cluster relationships.

While exploring intra-cluster relationships, Cheng et al. \cite{cheng2022hybrid} recognize that exclusively extracting positive examples from all memory instances might introduce inaccuracies owing to clustering imprecision. They consider the smaller number of clusters in each batch of data more reliable, as this reduced quantity should diminish the likelihood of selecting incorrect negative/positive sample pairs for contrastive learning. 

In \cite{cheng2022hybrid}, Cheng et al. address the potential inaccuracies in mining positive examples by introducing a batch-level contrastive loss, specifically operating between samples of the same batch. This is complemented by a standard contrastive loss that concurrently operates on all samples stored in memory. Additionally, they extend their approach to include unclustered examples, akin to \cite{ge2020self}. For each clustered sample, both losses designate the farthest example in its cluster as a positive instance: one loss retrieves this positive sample from the global memory bank, while the other does so at the local batch level. In the case of unclustered examples, the positive pair consistently resides in memory, representing its corresponding outlier.

\subsubsection{Feature Learning}
Feature Learning approaches in UDA for ReID share a common goal with their supervised counterparts, as discussed earlier: extracting discriminative features from raw images to accurately identify individuals. However, in contrast to the supervised scenario, feature learning in UDA is applied to unlabeled images, which introduces additional complexity. This increased difficulty arises due to the absence of ground-truth labels, making it challenging to guide the learning process effectively and requiring the model to discern relevant patterns without explicit supervision. 

In the most standard feature learning scenario, a feature extractor is employed to produce both local and global features: global features are extracted by operating on the entire feature maps, while local features are extracted by horizontal partitioning the feature maps into stripes. Ultimately, local and global features are employed as feature vectors for computing pseudo-labels through clustering methods. 

One of the early works that adopted this pipeline is by Fu et al. \cite{fu2019self}. Their framework utilizes ResNet50 as the CNN backbone, initially trained on source data. The same CNN backbone is then applied to generate features for target domain instances, yielding one global feature map and two local feature maps. The global feature aligns with the backbone-generated feature maps, while the two local features are obtained by horizontally partitioning these maps.

All the feature maps are grouped into three sets based on their type—distinctly categorizing global, upper-part local, and lower-part local features. Each set undergoes a  clustering algorithm, resulting in three labels for each sample corresponding to different feature types. Finally, triplet loss is computed for each set of feature maps, using the pseudo-labels generated by their respective clustering algorithms as labels.

This method treats each feature type independently, conducting clustering and computing loss separately for each set of features. Ding et al. \cite{ding2022learning} emphasize that this independent clustering on global and local features carries the risk of assigning multiple pseudo-labels to the same unlabeled sample.  Consequently, the model struggles to classify the sample to a specific identity during training, a phenomenon defined by the authors as "obscure learning."

To overcome this challenge, Ding et al. \cite{ding2022learning} propose the Learning Feature Fusion ($LF^2$) framework. This approach incorporates the Global-to-Local Fusion Module (FM), which seamlessly merges local and global features to generate corresponding pseudo-labels, aiming to mitigate the issue of obscure learning.

The $LF^2$ framework employs a mean teacher-student architecture to extract both global and local features. Specifically, the student model focuses on extracting local features by partitioning the extracted feature maps into two horizontal stripes, while the teacher network extracts the global feature maps. Both sets of features are then fed into the Fusion Module, where the global feature and one of the two local features are combined through element-wise multiplication. This process results in two fused feature vectors, each representing a distinct combination of local and global features. These fused feature vectors, along with the original global features, are utilized to predict pseudo-labels using a clustering algorithm, generating three pseudo-labels for each image. These pseudo-labels, together with soft labels of the global feature vectors, are used to compute soft triplet loss and a classification loss as in \cite{soft_labels_UDA}. 


Wang et al. \cite{wang2022learning} enhance the learning of global and local features using a memory-based contrastive approach. This is done by the Multiple Granularity Memory Dictionary Group (MGMDG) module, which employs contrastive learning for features at various granularities, each paired with a dedicated memory bank that stores one random feature vector for each pseudo-class. This system allows comparing the same detailed features with all the features extracted during the training, thereby elevating the robustness of the feature learning approach.

More precisely, they introduce a mean teacher-student framework that incorporates both source and target domain instances during training. In this configuration, identical samples are fed to both the student and teacher networks, extracting features at distinct granularities as horizontal stripes in the feature maps, each serving a unique purpose. The feature maps of the teacher network are concatenated in unique feature vectors for subsequent clustering, to avoid the obscure learning issue. The resulting pseudo-labels are then applied to the corresponding feature vectors of the student network. In the student network, the MGMDG module engages in contrastive learning, calculating the InfoNCE loss (\ref{eq:infoNCE}) for each granularity feature vector by mining samples from the associated memory bank. Finally, after each iteration, one random sample from each cluster in the batch is chosen to update the memory.

\subsubsection{Self-paced learning}
Self-paced learning (SPL) is a learning paradigm that adjusts the learning process by gradually introducing training samples based on their difficulty. SPL facilitates a more adaptive learning experience by allowing the model to focus on easy-to-learn samples initially and progressively introducing more complex instances. This approach is inspired by curriculum learning \cite{soviany2022curriculum}, a paradigm that mimics the natural learning process observed in humans, where the initial focus is on mastering easy samples before gradually incorporating more challenging ones into the learning journey. 

The SPL approach aligns with the intrinsic nature of person re-identification, where the appearance of individuals can vary significantly across different domains and scenarios. Especially in the unsupervised context of ReID, SPL provides a valuable means to steer the learning process in the absence of labelled data. This contributes significantly to alleviating the influence of noisy pseudo-labels, thereby augmenting the robustness and overall performance of the system.

In the most simple SPL techniques, the assessment of training sample difficulty is initially determined by a pre-existing condition of the samples. Subsequently, this condition is utilized to organize the samples in ascending order of difficulty, effectively guiding the training process.

Guo et al. \cite{guo2022gradual} exploit the different nature of source and target domain to guide the learning process. In particular, the authors employ a straightforward SPL technique that simultaneously trains its framework on both source and target domain instances. The rationale behind dividing the samples into simple and hard ones is to progressively assign greater importance to target domain instances while diminishing the significance of source domain ones. This gradual adjustment enables the model to shift its focus from initially mastering easier source samples, which benefit from ground truth labels, to gradually tackling more challenging target samples. The authors classify source samples as easy due to the provision of accurate labels, while target samples are considered harder due to their supervision by noisy and inaccurate clustering-based pseudo-labels. 

This SPL step dynamically adjusts the loss function based on domain-level instances, with the weight computed using a polynomial function in relation to the current epoch number. This approach leads to a systematic increase in weight for the target domain loss during training, concomitant with a decrease in the weight of the source domain loss.

Alternate SPL techniques assess the difficulty of training samples by employing clustering algorithms to organize samples into clusters based on their individual difficulty levels. The clustering algorithm is executed on samples, guided by an ad-hoc distance or score generated dynamically during the learning process.

Chen et al. \cite{chen2023mtnet} introduce a hard-sample filter that evaluates sample difficulty by employing the DBSCAN algorithm to detect the outlier samples, regarded as the hardest instances. Specifically, the Jaccard distance is used as a distance metric for clustering. To calculate this distance between two samples, a pairwise similarity matrix $M_{ij}$ is first computed. Subsequently, the Jaccard distance between samples $i$ and $j$ is determined by the following formula:
\[
\text{$Distance_{Jac}$}(i, j) = 1 - \frac{\sum_{k=1}^{N} \min(M_{ik}, M_{jk})}{\sum_{k=1}^{N} \max(M_{ik}, M_{jk})}
\]
where N is the number of samples. This distance essentially measures the dissimilarity between the sets represented by samples $i$ and $j$ where a lower Jaccard distance indicates higher similarity between the sets.

The DBSCAN clustering here categorizes unlabeled target samples into inliers and outliers. Inliers correspond to samples closer to the cluster center, featuring more accurate pseudo-labels. Consequently, the model is initially trained using these inlier samples, prioritizing their use. Outliers, characterized by increased difficulty, are incorporated into the training process at a later stage.

Liu et al \cite{liu2023dual}  assigned to each sample a reliability score, quantifying its cosine similarity with positive neighbours within the same cluster. Higher reliability indicates a stronger affinity with positive neighbours and a tendency to possess a clean pseudo-label. Subsequently, the K-means algorithm is applied to each cluster based on sample reliability, resulting in the division of clusters into three subsets: easy subset (D1), moderate subset (D2), and hard subset. Easier subsets, characterized by lower intra-class variance, are prioritized for learning as they are more likely to contain true-positive instances within the cluster. This sequential approach aims to prevent early model performance degradation by tackling noise first in the easier subset, followed by the moderate and hard subsets.
In \cite{zhao2022exploiting} Zhao et al presented another solution. In the first stage, the network is pretrained with source dataset in a supervised manner. In the second stage, they introduced an adaptive dynamic clustering (ADC) method for calculate $Eps$ of DBSCAN algorithm dynamically and a 
cross-camera similarity evalutation (CCSE) method for comparing and filter out noisy pseudo-labels caused by small inter-class variations under the same camera. 

\subsubsection{Final considerations and future directions}
In UDA for ReID, various approaches aim to improve adaptation from a labeled source to an unlabeled target domain. Clustering-based methods stand out as the best practice in this scenario. However, challenges persist in dealing with noisy pseudo-labels arising from domain gaps and clustering limitations. 

Mean teacher-student techniques introduce a dual-network paradigm to enhance pseudo-label quality, where the reliability of the teacher network is improved by accumulating knowledge over training iterations, leading to a more robust feature extraction process. Memory-based contrastive techniques use memory banks to retain feature vectors during training, allowing clustering on the complete dataset and mitigating the risk of noise in pseudo-labels. Feature learning approaches extract discriminative features from unlabeled images to guide the learning process. While effective, independent clustering on global and local features may assign multiple pseudo-labels to the same sample, causing challenges known as "obscure learning." SPL techniques adjust the learning process based on sample difficulty, either using pre-existing conditions or clustering algorithms.

In conclusion, UDA for ReID employs mean teacher-student, memory-based contrastive, feature learning, and self-paced learning techniques. Despite advancements, the primary challenge remains the presence of noisy pseudo-labels. Mean teacher-student and memory-based contrastive techniques enhance pseudo-label quality, while feature learning faces challenges of "obscure learning." Self-paced learning methods show promise in adapting the learning process based on sample difficulty. Future research should focus on enhancing the refinement of pseudo-label quality for more accurate and robust UDA models for ReID, addressing the persistent gap with supervised approaches.

\subsection{Fully Unsupervised learning}

Unsupervised approaches deal typically with unlabelled datasets where, conversely to the supervised methods, the unsupervised person re-identification problem revolves around learning to discriminate between individual images without any notion of semantic categories. Thus, given a training set 
\( X = \{x_1, \ldots, x_n\} \), 
the objective is to develop a model 
\( F(\theta, x_i) \)
where $\theta$ represents the parameters of the model and the function $F$ maps input images $x_i$
to their corresponding feature representations. A good model should map images of the same person closer to each other while pushing images of different identities further apart.

The inference phase typically follows the same approach as supervised techniques: ranking the distances between a query and all gallery images and the test is performed over a supervised dataset. 
\begin{figure*}[htb]
  \centering
  \includegraphics[width=0.7 \linewidth]{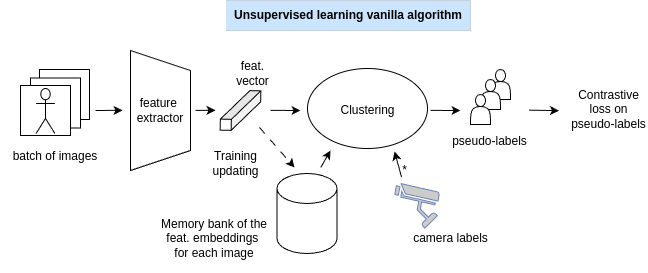}
  \caption{Vanilla model for unsupervised person ReID. The extracted features are stored in a dictionary for each image. These embeddings, combined with camera labels and a clustering algorithm, generate pseudo-clusters, which are then used to optimize contrastive losses.}
  \label{fig:unsupervised_vanilla}
\end{figure*}

The common baseline algorithm for training re-identification models leverages a combination of self-supervised clustering and contrastive learning techniques, schematically described in Figure \ref{fig:unsupervised_vanilla}. After extracting feature vectors from images using a backbone network, a clustering algorithm groups them based on a similarity metric. Each instance is then assigned to a cluster, and the centroid of that cluster is calculated as the mean of its embedded vectors. Subsequently, features from a batch of new images can be compared contrastively with these centroids, treating them as positive (similar) or negative (dissimilar) samples. 

The model maintains a memory bank storing feature embeddings for each image in the dataset. This helps in comparing current features with past features, enabling better clustering and pseudo-label generation for unsupervised learning. Camera labels and additional information, such as pose estimation, can be optionally included to enhance clustering by accounting for different perspectives.

However, this prevalent algorithm faces challenges due to several factors. In this section, we will delve into the key issues and challenges and present solutions explored in the literature.
Specifically, Section \ref{sec:funs-feature-learning} discusses the network architecture used for feature learning with unsupervised data.
The clustering methods and sampling strategies for enhancing intra-class and inter-class discrimination ability are eseprnted in Section \ref{sec:funs-clustering-sampling}.
Section \ref{sec:funs-noise-pseudolabels} discusses the problem 
created by noised pseudolabels during clustering, and some of the 
solutions proposed in literature.
Section \ref{sec:funs-instances-understanding} reviews solutions to understanding instance relationships in clustering. The section compare these techniques, noting their applications and limitations.
Finally, Section \ref{sec:funs-humancentrinc-perception} describes how advanced models for interpreting human perception from visual data can be used to address the re-identification task.

\subsubsection{Feature learning}
\label{sec:funs-feature-learning}
Choosing the appropriate backbone architecture for extracting feature vector at different granularities remain a recurrent problem also in the domain of unsupervised person re-identification learning.

In \cite{funs21} "LUPerson" (Large-scale Unlabeled Person Re-ID) dataset was presented. In the article Fu et al. proposed an algorithm to tackle unsupervised person re-identification. Similar to supervised methods, they employed a pre-trained ResNet50 backbone on ImageNet. To make the features suitable for contrastive learning, they added a two-layer MLP projection head on top of the ResNet50. Notably, their work utilized two separate encoders, one for the query image and another for the gallery set.
Similarly, the work of \cite{funs4} uses two branches for feature vector extraction. Each backbone has two different data augmentations. They showed that grayscale image transformation improves generalization when applied to one of the two branches.

In \cite{funs11}, a dual-branch architecture was appended to the output of a ViT. The images were tokenized with camera embeddings information. Then, the input was injected into a ViT as the backbone for feature extraction, and the output of the last layer was duplicated for two branches (local and global) to produce outputs independently. The local tokens in each branch were reshaped and uniformly partitioned into multiple stripes to yield partial-level features, while the global tokens of both branches were averaged to obtain a global feature. The training strategy employed was based on the 02CAP method, which will be described in the next part of this section.

Pre-training ViT models on LuPerson dataset can improve performance on benchmarks like Market and Duke. This was shown in \cite{funs20}. Significant differences between source and target distributions can lead ViT suffering of catastrophic forgetting problems leading to poor results. To mitigate this, the authors introduced a Catastrophic Degradation Score to assess distribution similarity and developed a filtering method that reduces the source dataset by 50\% of samples while maintaining benchmark performance. 

\subsubsection{Effective Clustering \& Sampling}
\label{sec:funs-clustering-sampling}
A major challenge in learning effective representations in the unsupervised setting involves the creation, organization, and sampling strategy of clusters. Without labels, it’s difficult to ensure that clusters accurately reflect individual identities. Incorrect clustering may group different identities together or split the same identity into multiple clusters. Deciding how to sample from clusters during training is critical. Poor sampling may result in inefficient learning, with the model focusing on irrelevant or redundant data.

The common approach across papers is to initialize a dictionary before training, using clustering results such as pseudo-labels, feature vectors of cluster centroids, outliers, or camera information, along with other relevant data. This dictionary is then updated at the end of each training epoch. In the literature, the labels assigned during clustering of embeddings are often referred to as pseudo-labels.

In \cite{intro_funs1}, a sampling strategy for contrastive learning is presented, involving the use of three memory banks. One is dedicated to global features, another to local features, and a combined memory bank concatenates the first two. \cite{funs1} promotes a memory based hard sample mining scheme to identify the hardest positive and negatives instances in the memory. This idea results into a combined loss that enhances both cluster-level contrastive loss and instance-level contrastive similarity.

Clustering based on camera information can potentially lead to better generalization learned from the model, as it enables learning of various dissimilarity factors among instances. In \cite{funs_UDA}, a stochastic method for sampling positive and negative instances is introduced. This technique advocates for the utilization of recently updated centroids of the cluster over others. Additionally, it incorporates camera information in the cluster construction.

The work \cite{funs7} utilizes real camera labels to group extracted features through a memory bank. In this approach, each mini-batch of data consists of image features, pseudo-labels, and real camera labels. Pseudo-labels are updated from the clustering at the beginning of each epoch. They employ the hard proxy instance sampling strategy, where instances with the minimum similarity to the proxy centroids are selected. This strategy is used for updating the camera proxy memory bank and for sampling intra-camera and inter-camera samples for the contrastive loss.

In \cite{funs6}, during each training iteration, camera centroids are computed for samples within the same cluster based on involved camera views to capture local structure. A time principle is then applied to select a distinct camera centroid as a proxy for each cluster, allowing the model to focus on the camera view causing the most intra-ID variances. Finally, all proxies are stored in a camera contrast proxy memory (CCPM).

In their work \cite{funs10}, Wang et al. introduced the O2CAP method, which divides each single camera-agnostic cluster into multiple camera-specific proxies to address the variance within clusters. They presented two types of inter-camera contrastive learning losses, named online and offline. The online loss is computed by sampling through the criteria of instance-proxy similarity and the nearest neighbor criterion based on camera information. On the other hand, for a given image $x_i$, its positive proxies from all cameras that share the same pseudo global label $y_i$ are identified. This association defines the inter-camera offline contrastive loss, wherein positive and negative proxies are retrieved across cameras.

Under similar considerations, the work in \cite{funs12} adopted a two-stage training strategy. The first training stage involved intra-camera learning, utilizing a distinct encoder for each camera. In the second stage, the focus shifted to inter-camera training to learn the global instance representation. The authors leveraged the results of intra-camera clustering as prior knowledge. This combination of training approaches proved to be very helpful in addressing another problem: the noise introduced by the model during the creation of pseudo-labels at training time.

To alleviate the learning issue of misclustered identities during training, \cite{funs16} proposed a method to generate support samples between neighboring clusters, aiming to associate samples of the same identity. A progressive linear interpolation strategy guides the support sample creation, and a label-preserving loss keeps these samples close to their original clusters.

\subsubsection{Learning from noised pseudo-labels}
\label{sec:funs-noise-pseudolabels}
In self-supervised clustering and contrastive learning, noise arises when incorrect labels or representations are acquired, leading to errors in clustering or similarity calculations and impacting the overall model performance. To mitigate this effect, relying solely on camera information is insufficient.

Recent works have employed a combination of techniques based on student-teacher and momentum updating strategies. In the context of unsupervised person re-identification, the momentum strategy involves updating the feature extraction memory according to the "speed" of their updates. This momentum serves to smooth out the updates to the feature extraction process, enabling it to adapt more gradually and potentially enhance convergence during training.
A momentum strategy is used in \cite{Cluster_Contrast} during training, updating the dictionaries of memorized pseudo centroid features with new values. The newly acquired feature information is partially merged with the existing feature information stored in the memory from the current batch. This approach aids in achieving consistency in feature representation across training iterations.

Also, the work of \cite{funs3} presents a strategy of training that involves momentum and the use of two functional modules applied in two separate student-teacher models: one module optimizes the feature extraction and one optimizes the label noised during the training. 
Notably, the student-teacher training approach was introduced in the previous sections.
In \cite{cheng2022hybrid}, to reduce noisy labels and achieve convergence, the projections of image features from both the student and teacher networks are pulled toward pseudo-class centers using sharpening techniques that minimize the probability regression objective.

Jaccard distance is used on pseudo label generation to reduce noise by measuring similarity through neighbor overlap. Recently, \cite{ca-jaccard} found that intra-camera samples often dominate the sets, leading to unreliable results by including irrelevant samples and excluding useful inter-camera ones. To address this, they propose the Camera Aware (CA) Jaccard Distance. First, a camera-aware k-reciprocal nearest neighbors (CKRNN) approach ensures that inter-camera samples are included in the top-k significant neighbors. Second, the Camera-Aware Local Query Expansion (CLQE) method extracts reliable samples from this set and assigns them higher weights to further enhance reliability.
The authors also presented versions of the baseline for both supervised (that use BoT \cite{luo2019bag}) and an UDA approaches (that use PPLR \cite{PPLR}), which we have included in the benchmarks.

Finally, \cite{funs18} address the noised label problem trough an inibition framework for estimating the confidence level of each pseudo label and actively correcting erroneous labels. Their method utilizes the silhouette coefficient to identify and recalibrate the Jaccard distance matrix in subsequent epochs. Essentially, the authors introduce the silhouette coefficient guided (SCG) InfoNCE loss to adjust the contribution of samples during model updating. Notably, silhouette the coefficient ranges between -1 and 1, with incorrect samples typically exhibiting a low silhouette coefficient close to -1.

\subsubsection{Instances Relationship understanding}
\label{sec:funs-instances-understanding}
Without supervision, it’s difficult to distinguish between intra-class variations (e.g., pose, lighting, or viewpoint changes for the same person) and inter-class similarities (e.g., different individuals with similar appearances).

A variety of approaches have been proposed for improving the understanding of the relationships between instances; some of them study the shape of the clusters, other try to catch the relationship in diffeent ways.

Early studies often assume that the output embeddings of images follow a spherical distribution, leading to the canonical approach of using momentum to update the memory dictionary. However, in practical scenarios, clusters may exhibit different embedding shapes, prompting the utilization of Euclidean distance and favoring methods like DBSCAN.

Instead of using momentum, \cite{funs17} promotes a real-time memory updating strategy 
where a stocastically sample $i-th$ dictionary entry $m_i$ is replaced by the current feature $f_i$ in the minibatch. 

The DiDal framework \cite{funs13} uses a graph to represent intra-class relationships among individuals. By organizing the embedding space in a graph-based structure, the model can understand both local and global relationships between identities. This structure allows for flexible handling of complex real-world challenges, like occlusions or varied poses, where individuals may appear differently across cameras. 

Attention mechanism was employed for the same purpose in \cite{funs19}. 
The authors constructed the embedding of the features by using an affinity/similarity matrix $R_i,j$ among all extracted features. Given a vector of the feature of an image, $x_i$, the rows of the matrix $r_i$ return the affinity to all the other extracted features and this allows to understand cluster correlation among the features.

In \cite{normalizedIBN}, the SimAM attention mechanism \cite{SimAM} is
used to dynamically allocate 3D attention weights to different areas, based on the input image content.


Most unsupervised domain adaptation methods generate one pseudo label for each unlabeled sample; this can be problematic, 
due to the complexity to capture a person accurately and 
may result in a large number of noisy labels by one-shot 
clustering. In \cite{funs15}, the authors propose to 
generate multiple pseudo labels to learn more diversified and complementary features through clustering. Clusters are then refined over time as the model learns more about the distinct features that differentiate individuals. as the number of pseudo-label increase, the entity of noise increase as well, so in the work, weights dictionaries are constructed to update the memorization during the training stage. Moreover, a student network is guided and aligned by
the teacher network, which resists noisy labels with the mean mechanism.

\subsubsection{Human-Centric Perception}
\label{sec:funs-humancentrinc-perception}
Recent research efforts \cite{class_1,UniHCP} have begun addressing person re-identification within the broader and more complex framework of Human-Centric Perception. This area of research focuses on understanding and interpreting human-related information from visual data, such as images and videos, with the ultimate goal of enabling machines to recognize, analyze, and predict various aspects of human presence, behavior, and interactions in ways that closely align with human understanding. Beyond person re-identification, Human-Centric Perception encompasses a range of tasks, including pose estimation, human parsing, pedestrian detection, and more. The underlying motivation for integrating these tasks is that they are inherently interconnected; the ultimate objective is to develop a unified model capable of learning and solving a wide array of human-centric tasks in a streamlined, end-to-end manner.

The UniHCP model \cite{UniHCP} exemplifies this approach by unifying a wide range of human-centric tasks using a plain vision transformer architecture. This model demonstrates how a single architecture can be effectively applied to various tasks within the human-centric perception domain.

Similarly, the Semantic cOntrollable seLf-supervIseD lEaRning framework (SOLIDER) \cite{class_1} leverages prior knowledge from human images to extract semantic information for various downstream tasks. SOLIDER generates pseudo labels for tokens and introduces a token-level semantic classification pretext task, which is supervised by these labels. Designed as a conditional network with a semantic controller, SOLIDER enables the dynamic adjustment of the semantic-to-appearance ratio in its representations, optimizing performance across different downstream tasks.

\subsubsection{Final considerations and future directions}
A fully unsupervised approach closely reflects real-world deployment scenarios, where the identities of individuals are typically unknown, making it a natural fit for production environments.

However, the absence of labeled data makes this task especially challenging, as the model must learn meaningful patterns and relationships without explicit supervision. This opens up several unresolved problems that require further investigation.

Pre-training on large datasets like LuPerson has emerged as a promising direction to enhance generalization. Such pre-training can work effectively with various feature extraction backbones, including ViT and ResNet50, improving performance in diverse settings.

One of the biggest challenges remains the organization of clustering and sampling strategies in pseudo-label generation. Pseudo-labels, when generated from unsupervised methods like clustering, tend to be noisy, leading to inconsistent or inaccurate learning. Momentum-based update strategies can smooth and stabilize training, but they may be less effective if clusters deviate from the assumed spherical shape. 
Effective methods for enhancing intra-class and inter-class discrimination usually exploit additional information like camera views and pose estimation.
Attention mechanisms and graph-based approaches offer alternative ways to learn meaningful relationships between instances.

Human-centric perception adds another level of complexity to Person Re-ID. The model needs to capture fine-grained human features and variations that span across different domains in a unified way, making this task highly demanding and multifaceted. Attention to domain-specific factors like pose, clothing changes, and camera variations will be essential in improving unsupervised Person Re-ID in practice.

\section{Benchmarks}
\label{sec:Experimental_result}
In this section, we will show and analyze the performance of the previously illustrated methodologies. Before delving into this discussion, it is important to introduce the datasets utilized and the metrics used for evaluation.

\subsection{Datasets}\label{sec:results_datasets}
In recent years, the re-identification (re-ID) community has collected several large datasets. These collections primarily consist of small images featuring a single individual, isolated through the use of bounding boxes. The edges of these boxes are sometimes manually outlined, while in other instances, they are detected automatically. In these collections, each image is labeled with information about the identity and camera from which it was captured, serving as ground truth labels. Additionally, some datasets include extra details like thermal imaging or sketches of the individuals. According to the most renowned benchmarks, model evaluation is conducted by dividing the data into training, testing (or gallery), and query sets. For each query — an image representing a specific identity — a cross-camera search is executed within the gallery (the test set).

In this study, we will introduce the datasets \textit{Market1501} and \textit{DukeMTMC}, as they are the two most widely employed datasets among the works discussed in this study, allowing for a fair comparison across various research contributions. \textit{Market1501} \cite{Scalable_person_re} was collected at Tsinghua University, covering 1,501 identities and 31,466 images.
Market1501 is divided into 750 identities for training set and 751 identities in test set. Query set contains test set images and benchmark evaluation protocols requires to count only image with different cameras.
The DukeMTMC dataset \cite{Duke} was collected at Duke University and it has 1,812 identities and 36,411 images. Both these datasets have been collected on campus, focusing on local space and short-term settings: the clothes of pedestrians do not change significantly. This assumption has a certain gap with real scenes and would limit the application in real-world scenarios. 
The dataset has 404 additional identities in the test set that are called distractors. Query set contains images test set identities. Conversely to Market1501, in the evaluation of mAP for DukeMTMC, the protocols do not impose any restrictions regarding cameras since there is no overlapping among camera networks. Further information can be found in Table \ref{table:4}. The dataset has been retracted by Duke University, although some works continue to use it. 


\begin{table*}[ht]
\centering
\resizebox{\textwidth}{!}{%
\begin{tabular}{|l|c|c|c|c|c|c|c|}
\hline
Dataset & Cams. & (HxWxC) & \#Images (IDs) & \#Training (IDs) & \#Gallery (IDs) & \#Query & Ref.\\
\hline
Market1501 & 6 & 128x64x3  & 32,668 (1,501) & 12,936 (750) & 19,732 (751)     & 3,368 & ICCV 2015  \\
DukeMTMC   & 8 & 256x128x3 & 36,411 (1,812) & 16,522 (702) & 17,661 (702+408) & 2,228 & ECCV 2016\\
\hline
\end{tabular}%
}
\caption{Market and Duke dataset information.}
\label{table:4}
\end{table*}

\subsection{Metrics}\label{sec:results_metrics}

The most used metrics for person re-identification are mAP and Rank-1 accuracy which ground truths can infer. 

For retrieval systems that provide a ranked sequence of documents, it is common practice to consider the order in which the returned documents are presented when evaluating their performance. For a given query $i$, precision and recall can be computed at each position in the ranked sequence of documents. By plotting precision against recall and calculating the area under this curve, we obtain the average precision for query $i$, denoted as $AP(i)$. The \textit{mean average precision} (mAP) is then simply the average of these individual average precisions across all possible queries. In more mathematical terms, mAP can be defined as follows:
 
\begin{equation}
    mAP = \frac{1}{N_Q}\sum_{i=1}^{N_Q}AP(i) = \frac{1}{N_Q}\sum_{i=1}^{N_Q}\sum_{i=1}^{N_G}[P_i(j) \cdot \Delta R_i(j)]
\end{equation}
where $P_i(j)$ is the (interpolated \cite{pascalVOC}) precision relative to the query $i$ at position $j$ (in the ranked list of retrievals), and $\Delta R_i(j)$ is the change in recall from position $j-1$ to position $j$.

\textit{Rank-1 accuracy (R1)} is a widely used performance metric in the field of person re-identification (person re-ID). It quantifies the effectiveness of a person re-ID system by assessing its ability to correctly match a query image to the most similar image of the same person in a gallery or dataset. The Rank-1 accuracy is calculated by summing up 1 for each query where the top-ranked candidate is indeed the correct match and then dividing by the total number of queries. This metric provides the percentage of correct matches when considering only the top-ranked candidates for each query. The R1 accuracy can be formulated as follows:

\begin{equation}
\text{Rank-1 Accuracy (R1)} = \frac{1}{N} \sum_{i=1}^{N} \delta(r_i, i)
\end{equation}

\begin{itemize}
\item \(N\) is the total number of queries.
\item \(i\) represents the index of a query.
\item \(r_i\) is the rank of the correct match for query \(i\). In other words, it is the position in the ranked list of candidates where the correct match is found for the query.
\item \(\delta(r_i, i)\) is the Kronecker delta function, which equals 1 if \(r_i\) is equal to \(i\) (indicating a correct match) and 0 otherwise.
\end{itemize} 

\subsection{Results for supervised person re-identification}\label{sec:results_supervised}
First, we are going to present and discuss the results of supervised techniques. Table 1 illustrates the best 10 publications on the Market1501 dataset presented in this survey, reporting also the performances on the Duke dataset. 

\begin{table*}[ht]
\centering
\begin{tabular}{| l | S | S | S | S | S | S | l }
\cline{2-5}
\multicolumn{1}{c|}{} & \multicolumn{2}{c|}{\textbf{Market1501}} & \multicolumn{2}{c|}{\textbf{ DukeMTMC}} &  \multicolumn{1}{c}{}  \\
\hline
\textbf{Paper} & \textbf{mAP} & \textbf{Rank-1}  & \textbf{mAP} & \textbf{Rank-1} & \textbf{Loss} \\
\hline
\cite{class_1} Solider & 93.9 & 96.9 & \text{-} & \text{-} & CL \\
\cite{sharma2021person} LA-Transformer & 94.5 & 98.3 & \text{-} & \text{-} & CL \\
\hline
\cite{contr_1} VRN & 95.2 & 96.3 & \text{-} & \text{-} & CO  \\
\hline
\cite{MSINet} MSINet & 89.6 & 95.3  & \text{-} &\text{-} & CLTR \\
\cite{he2021transreid} TransReID & 89.5 & 95.2  & 82.1 & 91.1 & CLTR \\
\cite{wieczorek2021unreasonable} Wieczorek & 98.3 & 98.0 & 92.7 & 94.5 & CLTR \\
\cite{class_2} stReid & 95.5 & 98.0 & 92.7 & 94.5 & CLTR \\
\cite{class_tripl_4} LightMBN & 91.5 & 96.3 & \text{-} & \text{-} & CLTR \\
\cite{flipreid} FlipReID & 94.7 & 95.8 & 90.7 & 93.0 & CLTR \\
\cite{class_tripl_5} LDS & 94.9 & 96.2 & 92.9 & 91.0 & CLTR \\
\cite{alnissany2023modified} Mctl & \textbf{98.6} & \textbf{98.4} & \textbf{97.3} & \textbf{96.8} & CLTR \\
\cite{li2023clipreid} Clip-ReID & 90.5 & 95.4 & 83.1 & 90.8 & CLTR \\

\cite{ca-jaccard} CA-Jaccard & 94.5 & 96.2 & \text{-} & \text{-} & CLTR \\
\hline
\end{tabular}
\caption{Performance of best-supervised person re-identification publications. If the performance of a dataset is not reported in the original work, we use the symbol '-'. On the loss column, CL stands for classification loss, 'TR' for triplet, 'CO' for contrastive, and 'CLTR' for the combination of classification and triplet loss. The best work is highlighted in bold.}
\label{table:1}
\end{table*}

\begin{figure*}[htb]
  \centering
  \includegraphics[width=1.0 \linewidth]{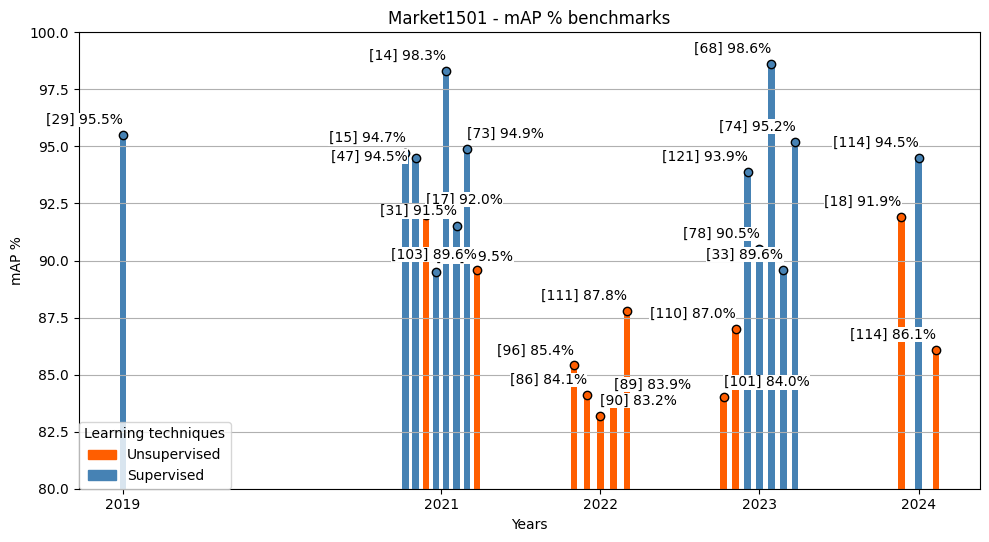}
  \caption{This figure illustrates the mean Average Precision (mAP) of the Market1501 dataset for both supervised and unsupervised techniques among years. The results highlight the performance comparison between these two categories of methods. Note that DukeMTMC benchmarks are not included, as many models were not evaluated on this dataset.}
  \label{fig:market-benchmarks}
\end{figure*}

Notably, papers employing the combination of triplet and classification loss demonstrate stronger performance across both datasets, surpassing other methods. Globally, all these works exhibit mAP well above $90\%$, with almost all hovering around $90\%$. Compared to Market1501, the available performances on DukeMTMC are slightly lower but still above $90\%$.  Specifically, \cite{alnissany2023modified} stands out, achieving the top-tier results on both Market1501 and DukeMTMC, leaving minimal room for further improvement.

\subsection{Results for unsupervised person re-identification}\label{sec:results_unsupervised}
In the realm of unsupervised person re-identification, it is noteworthy to initially compare the two presented UDA techniques before delving into a comparison with fully unsupervised methods. However, the evaluation of Domain Translation methods introduces a challenge, as it involves different datasets as source and target domains, complicating a fair comparison. Moreover, their performances consistently fall short when compared to UDA Clustering-based methods. Given these considerations, we opt for a direct comparison between the top five publications in UDA clustering-based methods and the top five publications in fully supervised methods on the Market1501 dataset, reporting also performances on DukeMTMC. This evaluation is based on their performances on Market1501 and DukeMTMC, as detailed in Table 2. Notably, for UDA clustering-based works, the 'Market1501' column reflects the dataset as the target and DukeMTMC as the source, while the 'DukeMTMC' column represents the reverse scenario (DukeMTMC as the target and Market1501 as the source).

\begin{table*}[ht]
\centering
\begin{tabular}{|l|S|S|S|S|S|S|l|}
\cline{2-5}
\multicolumn{1}{c|}{} & \multicolumn{2}{c|}{\textbf{Market1501}} & \multicolumn{2}{c|}{\textbf{DukeMTMC}} & \multicolumn{1}{c}{} \\
\hline
\textbf{Paper} & \textbf{mAP} & \textbf{Rank-1} & \textbf{mAP} & \textbf{Rank-1} & \textbf{Technique} \\
\hline
\cite{zhao2022exploiting} ZhaoRN50 & 84.1 & 92.9 & 74.8 & 86.1 & \text{UDA}C \\
\cite{song2022dual} DPCFG & 85.4 & 94.2 & 73.7 & 85.7 & \text{UDA}C \\
\cite{cheng2022hybrid} HDCRL & 83.9 & 93.6 & 70.9 & 83.5 & \text{UDA}C \\
\cite{ding2022learning} LF$^{2}$ & 83.2  & 92.8 & 73.5 & 83.7 & \text{UDA}C \\
\cite{liu2023dual} DUCL+PAFR & 84.0 & 93.9 & 72.7 & 84.9 & \text{UDA}C \\
\cite{ca-jaccard} CA-Jaccard & 86.1	& 94.4 & \text{-} & \text{-}& \text{UDA}C\\
\hline
\cite{funs12} Repel+Absorb & 87.0  & 94.5 & 80.3  & 89.9 & FUNS \\
\cite{funs11} TMGF(w/GT)$\dagger$ & 91.9 & 96.3 & 83.1 & \textbf{92.3} & FUNS \\
\cite{funs16} ISE & 87.8  & 95.6 & \text{-} & \text{-} & FUNS \\
\cite{funs21} LupRN50 & \textbf{92.0}  & \textbf{97.0} & \textbf{84.1} & 91.9 & FUNS \\
\cite{funs20} CFS & 89.6 & 95.3 & \text{-} & \text{-} & FUNS \\

\hline
\end{tabular}
\caption{Performance of top-5 unsupervised person re-identification publications both for UDA and fully unsupervised techniques. If the performance of a dataset is not reported in the original work, we use the symbol '-'. On the technique column, 'UDAC' stands for UDA clustering, and 'FUNS' for fully unsupervised. The best work is highlighted in bold.}
\label{table:2}
\end{table*}

This table highlights the robustness of UDA clustering techniques for each dataset. In fact, on the Market1501 dataset, the mAP performances are consistently around $85\%$, while on the DukeMTMC dataset, they range between $70\%$ and $75\%$. However, the gap between the two datasets is quite evident, reaching a difference of at least $10\%$ for each work. Regarding fully unsupervised techniques, here too, the performances consistently exceed $85\%$ on the Market1501 dataset, while the lack of data on the DukeMTMC dataset prevents a direct comparison. Finally, when comparing fully unsupervised techniques with UDA clustering techniques, we notice that the former consistently exhibit higher performances, with \cite{funs21} showing the best results thanks to a mAP of $92\%$. 

\subsection{Comparison between supervised  and unsupervised person re-identification}\label{sec:results_comparison}
Finally, we compare the top 5 supervised and unsupervised works. The outcomes are detailed in Table 3.

\begin{table*}[ht]
\centering
\begin{tabular}{|l|S|S|S|S|S|S|l|}
\cline{2-5}
\multicolumn{1}{c|}{} & \multicolumn{2}{c|}{\textbf{Market1501}} & \multicolumn{2}{c|}{\textbf{DukeMTMC}} & \multicolumn{1}{c}{} \\
\hline
\textbf{Paper} & \textbf{mAP} & \textbf{Rank-1} & \textbf{mAP} & \textbf{Rank-1} & \textbf{Technique} \\
\hline
\cite{wieczorek2021unreasonable} Wieczorek & 98.3 & 98.0 & 96.1 & 95.6 & SL \\
\cite{contr_1} VRN & 95.2 & 96.3 & \text{-} & \text{-} & SL  \\
\cite{class_2} stReid & 95.5 & 98.0 & 92.7 & 94.5 & SL \\
\cite{class_tripl_5} LDS & 94.9 & 96.2 & 92.9 & 91.0 & SL \\
\cite{alnissany2023modified} Mctl & \textbf{98.6} & \textbf{98.4} & \textbf{97.3} & \textbf{96.8} & SL \\
\hline
\cite{funs12} Repel+Absorb & 87.0  & 94.5 & 80.3  & 89.9 & USL \\
\cite{funs11} TMGF(w/GT)$\dagger$ & 91.9 & 96.3 & 83.1 & \textbf{92.3} & USL \\
\cite{funs16} ISE & 87.8  & 95.6 & \text{-} & \text{-} & USL \\
\cite{funs21} LupRN50 & \textbf{92.0}  & \textbf{97.0} & \textbf{84.1} & 91.9 & USL \\
\cite{funs20} CFS & 89.6 & 95.3 & \text{-} & \text{-} & USL \\
\cite{zhao2022exploiting} ZhaoRN50 & 84.1 & 92.9 & 74.8 & 86.1 & USL \\
\cite{song2022dual} DPCFG & 85.4 & 94.2 & 73.7 & 85.7 & USL \\
\hline
\end{tabular}

\caption{Performance of top-5 supervised and unsupervised person re-identification publications on both datasets. If the performance of a dataset is not reported in the original work, we use the symbol '-'. On the technique column, 'SL' stands for supervised learning, and 'USL' for unsupervised learning. The best works for each paradigm are highlighted in bold.}
\label{table:3}
\end{table*}

As expected, supervised techniques demonstrate superior performance on both datasets compared to their unsupervised counterparts, with \cite{alnissany2023modified} emerging as the top-performing work once again. Supervised methods consistently achieve a mAP on the Market1501 dataset of approximately $95\%$, with \cite{alnissany2023modified, wieczorek2021unreasonable} surpassing $98\%$ on Market1501. It is noteworthy that \cite{alnissany2023modified} also improves performance on DukeMTMC, reaching a mAP of $97\%$. These significant results highlight the near saturation of performance for supervised techniques on both datasets, leaving very little room for improvement.

Unsupervised approaches, on the other hand, range from $86\%$ to $90\%$ on the Market1501 dataset, except for \cite{funs21} and \cite{funs11}, which reach $92\%$. The observed gap between the top supervised and top unsupervised methods is about $6\%$ on the Market1501 dataset. This difference is not far from being considered acceptable, especially when taking into account the absence of labelled data in training and achieving mAPs around $92\%$. Furthermore, these unsupervised works, unlike supervised ones, still have room for improvement. The scenario differs when we compare results on DukeMTMC, where the gap with supervised techniques is still significant (approximately $13\%$).

\section{Conclusion}
\label{sec:conclusions}
In this work, we have compiled and analyzed state-of-the-art methods for both supervised and unsupervised person re-identification. For the supervised approaches, we categorized the methods based on the loss functions they utilize, and our comparison highlights that the combination of classification and triplet loss consistently yields the best results. For the unsupervised approaches, we collected the top-performing methods from the past three years and classified them into two categories: Unsupervised Domain Adaptation and Fully Unsupervised methods. Our comparison indicates the superiority of fully unsupervised approaches in this domain.

We then conducted a comparative analysis of the performance of both learning paradigms on two widely used person re-identification datasets: Market1501 and DukeMTMC. The state-of-the-art supervised methods demonstrate outstanding performance on both datasets, achieving around 98\%, thereby indicating that further improvements in this area are likely to be minimal. In contrast, state-of-the-art unsupervised methods show varying levels of performance depending on the dataset. On the Market1501 dataset, the performance gap between the two paradigms is approximately 6\%, which is relatively small, especially considering that unsupervised methods do not rely on labeled data and still achieve around 92\% performance. However, on the DukeMTMC dataset, the performance gap is more pronounced, amounting to 13\%.

In summary, our analysis demonstrates the near saturation of supervised performance on both the Market1501 and DukeMTMC datasets. Additionally, we highlight the commendable performance of unsupervised techniques on the Market1501 dataset, where the gap with supervised methods is relatively narrow. Conversely, there remains a substantial performance gap between supervised and unsupervised methods on the DukeMTMC dataset.

Despite recent advancements in unsupervised person re-identification (ReID), developing effective methods that can fully exploit unlabelled data while simultaneously learning robust and discriminative features for person identification remains a significant and ongoing challenge. The absence of labeled data necessitates innovative approaches to ensure that the learned features are both distinctive and invariant to the wide range of variations encountered in real-world scenarios.

One of the primary strategies employed by unsupervised methods is the use of clustering algorithms to group similar person images together. However, the effectiveness of this approach is often hampered by several factors, including noise in the data, variations in pose, lighting, and viewpoint, as well as the inherent ambiguities in person appearances. These challenges can cause clustering algorithms to struggle with forming coherent and accurate groups, leading to less reliable pseudo-labels for subsequent learning stages. As a result, the enhancement of clustering techniques to produce more precise and dependable clusters remains a largely unsolved problem, crucial for advancing the effectiveness of unsupervised ReID systems.

Among the challenges facing unsupervised person ReID, domain adaptation is perhaps the most critical and persistent issue. Models trained on a specific dataset typically exhibit a marked decline in performance when applied to a different domain, a problem exacerbated by variations in environmental factors such as lighting conditions, camera angles, background clutter, and sensor characteristics. These domain shifts can drastically alter the appearance of the same person across different camera views, making it difficult for the model to generalize effectively.

The core challenge lies in developing models that can generalize across diverse domains without the need for labeled data from the target domain. This necessitates the improvement of domain adaptation techniques to handle the complex and often unpredictable variations in appearance and environmental conditions. Current approaches, while promising, still fall short of fully addressing the intricacies involved in domain adaptation for unsupervised person ReID. The goal is to create models that are robust to these variations, enabling consistent performance across different domains without relying on costly and labor-intensive labeled datasets.

Another topic that is likely to play a growing role in coming years
is related to data security and privacy concerns. With the implementation of data protection regulations such as the General Data Protection Regulation (GDPR) in Europe and similar laws in other countries, there is increasing legal pressure to ensure that ReID systems comply with privacy standards. The task of anonimization is e.g. discussed in \cite{SecureReId}, stressing the
delicate balance between privacy requirements, information loss, and
robustness to adversarial attacks. 

An interesting perspective in this field is offered by neuromorphic vision sensors, also known as
event cameras. These sensors are distinct from conventional RGB cameras in that they only capture changes in brightness within the scene, rather than detailed visual images of individuals, inherently offering a greater degree of privacy by design.
The potentiality and limitations of this approach are discussed
in \cite{EventAnonymization}.

\subsection*{Acknowledgements}

This research was conducted in the framework of the Future AI Research (FAIR) project of the National Recovery and Resilience Plan (NRRP), Mission 4 Component 2 Investment 1.3 funded from the European Union - NextGenerationEU.

\subsection*{Statements and Declarations}
\begin{itemize}
\item The authors declare no competing interests;
\item all authors contributed equally to this work;
\item the article used public datasets and no ethical and informed consent for data was required.
\end{itemize}

\bibliography{references.bib}

\end{document}